\documentclass[letterpaper]{article} 
\usepackage{aaai2026}  
\usepackage{times}  
\usepackage{helvet}  
\usepackage{courier}  
\usepackage[hyphens]{url}  
\usepackage{graphicx} 
\usepackage{natbib}  
\usepackage{caption} 
\usepackage{algorithm}
\usepackage{algorithmic}
\usepackage[table]{xcolor}
\usepackage{colortbl}
\usepackage{amsmath}
\usepackage{amssymb}
\usepackage{graphicx}
\usepackage{booktabs}
\usepackage{siunitx} 
\usepackage{makecell} 
\usepackage{multirow}

\usepackage{newfloat}
\usepackage{listings}
\DeclareCaptionStyle{ruled}{labelfont=normalfont,labelsep=colon,strut=off} 
\floatstyle{ruled}
\newfloat{listing}{tb}{lst}{}
\floatname{listing}{Listing}
\usepackage{amsmath}
\usepackage{amsfonts}
\usepackage{booktabs}
\usepackage{subcaption}
\usepackage{array}
\usepackage[dvipsnames]{xcolor}
\title{MDBench: Benchmarking Data-Driven Methods for Model Discovery}
\author {
    Amirmohammad Ziaei Bideh\textsuperscript{\rm 1},
    Aleksandra Georgievska\textsuperscript{\rm 2},
    Jonathan Gryak\textsuperscript{\rm 1,2}
}
\affiliations {
    \textsuperscript{\rm 1}Department of Computer Science, Graduate Center, CUNY, New York, NY, USA\\
    \textsuperscript{\rm 2}Department of Computer Science, Queens College, CUNY, New York, NY, USA \\
    aziaeibideh@gradcenter.cuny.edu, aleksgeorgi@gmail.com, jonathan.gryak@qc.cuny.edu
}

\begin{document}

%
%
\maketitle

\begin{abstract}
Model discovery aims to uncover governing differential equations of dynamical systems directly from experimental data. Benchmarking such methods is essential for tracking progress and understanding trade-offs in the field. While prior efforts have focused mostly on identifying single equations, typically framed as symbolic regression, there remains a lack of comprehensive benchmarks for discovering dynamical models. To address this, we introduce MDBench, an open-source benchmarking framework for evaluating model discovery methods on dynamical systems. MDBench assesses 12 algorithms on 14 partial differential equations (PDEs) and 63 ordinary differential equations (ODEs) under varying levels of noise. Evaluation metrics include derivative prediction accuracy, model complexity, and equation fidelity. We also introduce seven challenging PDE systems from fluid dynamics and thermodynamics, revealing key limitations in current methods. Our findings illustrate that linear methods and genetic programming methods achieve the lowest prediction error for PDEs and ODEs, respectively. Moreover, linear models are in general more robust against noise. MDBench accelerates the advancement of model discovery methods by offering a rigorous, extensible benchmarking framework and a rich, diverse collection of dynamical system datasets, enabling systematic evaluation, comparison, and improvement of equation accuracy and robustness.

\end{abstract}

\begin{links}
    \link{Code}{https://github.com/gryaklab/mdbench}
    \link{Datasets}{https://zenodo.org/records/17611099}
    \link{Extended version}{https://arxiv.org/abs/2509.20529}
\end{links}
\section{Introduction}

Data-driven equation discovery---rather than relying solely on conservation laws or physical principles---has historically played a pivotal role in scientific breakthroughs. Johannes Kepler discovered the third law of planetary motion by leveraging geometrical intuition and searching for patterns in empirically gathered data \cite{cranmer2023interpretable}. Similarly, Edwin Hubble identified the relationship between redshift and distance through empirical observations that, at the time, lacked theoretical explanation \cite{kragh2021cosmology}.

A dynamical system is a mathematical framework that describes how the state of a system evolves over time through differential equations. Many complex real-world systems arising in finance, medicine, and engineering can be modeled as nonlinear dynamical systems, e.g., in physics, they model fluid dynamics \cite{kleinstreuer2018modern}; in biology, they describe neural activity \cite{breakspear2017dynamic}; and in engineering, they underpin control theory \cite{walker2013dynamical}.

\begin{figure}[t]
    \centering
    \includegraphics[width=1\linewidth]{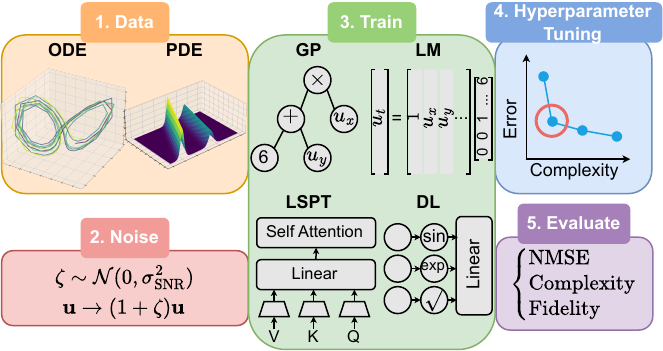}
    \caption{A schematic overview of MDBench pipeline.}
    \label{fig:mdbench}
\end{figure}

Relying solely on expert domain knowledge to derive governing equations is increasingly impractical, especially in the era of high-dimensional and large-scale experimental data. This has motivated the development of automated tools for scientific discovery. While machine learning (ML) models have demonstrated remarkable predictive performance, their black-box nature often impedes interpretability and insight into the underlying system dynamics. Data-driven model discovery (MD) aims to bridge this gap by using measurement data and ML algorithms to infer interpretable dynamical system models.

Unlike traditional black-box ML methods, MD involves optimizing both the parameters and the structure of the model, resulting in models that are inherently interpretable. However, generic MD techniques are not directly applicable to the discovery of differential equations. In this work, we present an extension of MD tailored specifically to the discovery of dynamical system models. In the literature, symbolic regression (SR) typically refers to the discovery of a single equation rather than a system. Here, we adopt the term model discovery to reflect the broader scope of our work.

Despite recent progress, the MD field lacks a standardized benchmark for evaluating algorithms on dynamical systems. To address this, we introduce \textbf{MDBench}, an open-source and extensible benchmarking framework, that includes a suite of ODE and PDE datasets (Figure \ref{fig:mdbench}). Establishing such a benchmark is essential for characterizing algorithmic strengths and weaknesses, facilitating fair comparison, and tracking the evolution of the field over time. We encourage researchers to contribute their methods and datasets to MDBench, enabling more robust and reproducible performance evaluations in terms of predictive accuracy, equation complexity, and robustness to noise. Our main contributions are as follows:

\begin{enumerate}
    \item We introduce the first comprehensive testbed for model discovery, covering a diverse set of methods, including linear models; genetic programming; deep learning; and large scale pretraining approaches, and spanning both ODE and PDE systems.
    \item We contribute new datasets simulating real-world PDE systems with up to six state variables, providing significantly more challenging benchmarks than existing MD datasets.
    \item We conduct systematic evaluations of state-of-the-art MD methods on these datasets, assessing their performance in equation re-discovery, predictive accuracy, and model complexity under varying noise conditions.
\end{enumerate}

The remainder of this paper is organized as follows. Section~\ref{sec:related_work} reviews existing MD algorithms and related benchmarks. Section~\ref{sec:mdbench} details the structure of MDBench, including datasets, training pipeline, metrics, and evaluated methods. Section~\ref{sec:results} presents experimental results and highlights current limitations. Finally, Section~\ref{sec:conclusion} concludes with a summary of our work. 
\section{Related Work}
\label{sec:related_work}


This section reviews prior benchmarking efforts in symbolic and scientific model discovery, identifies their limitations, and categorizes the landscape of existing MD methods to four classes.


\subsection{Overview of Benchmarks}

Several efforts have been made to benchmark MD algorithms. One of the earliest large-scale initiatives is SRBench \cite{la2021contemporary}, which introduced a publicly available benchmark evaluating 14 SR methods on a variety of regression problems. A later extension, SRBench 2.0 \cite{aldeia2025call}, expanded the number of evaluated algorithms to 25 and introduced additional metrics such as energy consumption. Notably, the authors concluded that no single SR method consistently outperforms others across all metrics, and they proposed several best practices for future method development such as minimizing hyperparameters and standardizing evaluation procedures.

Despite these contributions, these benchmarks are limited to time-invariant regression problems involving a single mathematical equation and do not account for dynamical systems governed by differential equations. Model discovery for dynamical systems presents additional challenges such as handling systems of coupled equations, high dimensionality, and numerical differentiation.

PDEBench \cite{takamoto2022pdebench} is a more recent benchmark focused on PDE systems that evaluated 11 physical systems using four machine learning surrogates: U-Net \cite{ronneberger2015u}, Fourier Neural Operator \cite{li2021fourier}, and Physics-Informed Neural Networks (PINNs) \cite{raissi2019physics}. While these methods are effective for solving PDEs, they are black-box models and cannot generate symbolic equations, limiting their utility for interpretable model discovery.

Gilpin et al. \cite{gilpin2021chaos} benchmarked 131 chaotic ODE systems, framing them as time series forecasting tasks. Although their study includes symbolic regression methods, only four MD methods are tested, and no evaluation is provided on the complexity or interpretability of the discovered models.

Other benchmarks have narrower or domain-specific scopes. cp3-bench \cite{thing2025cp3} evaluated 12 MD algorithms on 28 datasets from cosmology and astroparticle physics. CFDBench \cite{luo2023cfdbench} focuses on computational fluid dynamics (CFD) simulations and evaluated neural operators under varying conditions, without considering symbolic recovery of governing equations. The benchmark by Otness et al. \cite{otness2021extensible} includes four PDE systems, but evaluated only black-box predictive models such as convolutional neural networks and $k$-nearest neighbors.

Table \ref{tab:benchmark_comparison} summarizes key characteristics of these benchmarks, highlighting MDBench's broader coverage across ODEs, PDEs, and model discovery algorithms.

\begin{table}[ht]
\centering
\begin{tabular}{lcccc}
\toprule
\textbf{Benchmark} & \textbf{\#ODE} & \textbf{\#PDE} & \textbf{\#MD} & \textbf{Noise} \\
\midrule
SRBench 1.0 & 7 & 0 & 14 & \checkmark \\
SRBench 2.0 & 0 & 0 & 25 & \checkmark \\
PDEBench & 0 & 11 & 0 & \\
\cite{gilpin2021chaos} & 131 & 0 & 4 & \\
cp3-bench& 7 & 0 & 12 & \checkmark \\
CFDBench & 0 & 4 & 0 & \\
\cite{otness2021extensible} & 0 & 4 & 0 & \\
\textbf{MDBench (ours)} & 63 & 14 & 12 & \checkmark \\
\bottomrule
\end{tabular}
\caption{Comparison of existing benchmarks for scientific MD. \textbf{\#ODE} and \textbf{\#PDE} refer to the number of ODE and PDE problems, \textbf{\#MD} refers to the number of model discovery algorithms evaluated, and \textbf{Noise} indicates whether robustness to noise is studied.}
\label{tab:benchmark_comparison}
\end{table}

\vspace{-1em}

\subsection{Overview of Model Discovery Methods}

Since Koza's early work on genetic programming (GP) for symbolic regression \cite{koza1994genetic}, numerous methods have emerged for data-driven model discovery. We categorize them into four main classes.

\subsubsection{Genetic Programming (GP)}

GP-based methods evolve expression tree of equations using evolutionary operators to search for equations that best fit the data. Modern and efficient implementations, such as PySR \cite{la2021contemporary} and Operon \cite{burlacu2020operon}, are widely used in the literature.

GP methods often struggle with convergence due to their large, unstructured search spaces. They also do not naturally learn parameters from data. Neural-guided approaches like those in \cite{mundhenk2021symbolic} mitigate this by incorporating learned search heuristics. Recent work further accelerates GP convergence by using large language models (LLMs) as crossover and mutation operators \cite{shojaee2025llm, meyerson2024language}.

\subsubsection{Linear Models (LM)}

Linear model-based methods represent equations as sparse linear combinations of candidate basis functions (``library"). A typical form is $f = \sum_{i=1}^{k} \beta_i \phi_i$, where $\phi_i$ are predefined basis terms and $\beta_i$ are scalar coefficients. SINDy \cite{brunton2016discovering} pioneered this approach using LASSO for sparsity. PDEFIND \cite{rudy2017data} extended it to PDEs, and WSINDy \cite{messenger2021weak} further improved robustness by formulating equations in weak form, avoiding direct differentiation. ESINDy \cite{fasel2022ensemble} introduced ensemble learning for robustness in low-data or high-noise regimes. DeepMoD \cite{both2021deepmod} combines neural networks with sparse regression by learning spatial features via automatic differentiation.

Bayesian variants \cite{yuan2023machine, north2025bayesian, more2023bayesian} introduce uncertainty modeling and noise-aware inference. However, all LM methods are constrained by the assumption of linearity with respect to basis functions and often lack principled ways to construct these function libraries for unknown systems.

\subsubsection{Large-Scale Pretraining (LSPT)}
LSPT methods pretrain a single model---often a Transformer architecture \cite{vaswani2017attention}---on a large corpus of symbolic regression problems. These models learn to map input–output pairs to symbolic equations, enabling rapid inference on new data once training is complete.

Neural Symbolic Regression that Scales (NeSymReS) \cite{biggio2021neural} exemplifies this approach. It uses a Set Transformer \cite{pmlr-v97-lee19d} as an encoder to learn latent representations of input–output mappings from synthetic equations. A decoder then generates symbolic expressions autoregressively using beam search. After generating symbolic forms, constant placeholders are refined using post-hoc optimization with methods such as BFGS \cite{fletcher2000practical}.

To overcome limitations of this two-stage design, Kamienny et al. \cite{kamienny2022end} introduced an end-to-end approach where the Transformer directly generates complete mathematical expressions, including constants. ODEFormer \cite{d2024odeformer} further extends this framework to dynamical systems, generating full differential equations from a single observed trajectory.

The main advantage of LSPT methods is their inference speed. Once trained, they can quickly generate symbolic equations for new datasets without the need for retraining or optimization, which makes them suitable for real-time applications. However, they may perform poorly on real-world systems whose dynamics differ from the synthetic equations seen during pretraining \cite{kamienny2022end}.

\subsubsection{Deep Learning (DL)} DL-based methods use neural networks to learn symbolic equations from data. One example is Equation Learner Networks (EQL) \cite{martius2016extrapolation, sahoo2018learning}, which are fully differentiable feed-forward networks that incorporate symbolic operators (e.g., sine, cosine, multiplication) as activation functions. This design enables the integration of symbolic structure into neural models and allows EQL to be combined with other architectures for scientific discovery tasks \cite{kim2020integration}.

Another class of DL methods uses recurrent neural networks (RNNs) to learn a probability distribution over symbolic expressions conditioned on data. Deep Symbolic Regression (DSR) \cite{petersen2021deep} follows this approach. It trains an RNN using a risk-seeking policy gradient algorithm, where the reward is based on the predictive accuracy of sampled equations. However, the learning signal in this setup comes only from the scalar reward based on the $(X, y)$ fit, which can be weak and indirect.

To overcome this limitation, uDSR \cite{landajuela2022unified} extends DSR by integrating optimization techniques from GP, LM, and LSPT approaches as inner loops. This hybrid approach achieves state-of-the-art performance on the SRBench benchmark \cite{la2021contemporary}.

\section{MDBench}
\label{sec:mdbench}

MDBench is a unified benchmarking framework for evaluating data-driven MD algorithms on both ODE and PDE systems (Figure \ref{fig:mdbench}). It includes a diverse suite of 63 ODEs and 14 PDEs, ranging from simple linear dynamics to high-dimensional physical systems. MDBench standardizes data formats, provides symbolic preprocessing for PDEs, incorporates realistic noise modeling, and supports automated hyperparameter tuning across methods. 

\subsection{Datasets}

Dynamical systems are commonly modeled using either ODEs or PDEs, depending on whether the variables of interest evolve only over time or over both time and space. ODE-based systems involve temporal dynamics, while PDE-based systems also incorporate spatial variation.

\subsubsection{ODE Systems.} 
ODEs describe the evolution of a system's state variables over time. Formally, a dynamical system with $d$ state variables $\mathbf{u}=(u_1,\ldots,u_d) \in \mathbb{R}^d$ is governed by equations of the form:
\[
\frac{du_i}{dt} = f_i(t,u_1, u_2, \dots, u_d), \quad i = 1, \dots, d,
\]
where $f_i: \mathbb{R}^d \rightarrow \mathbb{R}$. Each dataset consists of observed trajectories $\mathbf{U} \in \mathbb{R}^{N_t \times d}$, where $N_t$ is the number of samples across time.

We adopt the \textbf{ODEBench} dataset~\cite{d2024odeformer}, which includes 63 systems from a textbook by Strogatz~\cite{strogatz2024nonlinear} and several sourced from Wikipedia. These systems span one to four state variables and cover a range of real-world phenomena. For all methods, the input consists of state variables $\mathbf{u}$, and the targets are their time derivatives $\dot{\mathbf{u}}$.

\subsubsection{PDE Systems.}

PDEs describe spatiotemporal systems where each state variable depends on both space and time. A PDE system with $d$ state variables $\mathbf{u}=(u_1,\ldots,u_d) \in \mathbb{R}^d$ is represented on a $D$-dimensional spatial grid  $\mathbf{X} \in \mathbb{R}^{N_{x_1} \times \cdots \times N_{x_D}}$. The spatial grid is constructed from uniformly-spaced intervals along each spatial coordinate, with spacing that may differ between them (i.e., $\Delta x_1 \neq \cdots \neq \Delta x_D)$.  The time domain $t \in [0, T]$ is uniformly spaced. 

We focus on PDEs of the form:
\begin{equation}
    \frac{\partial u_i}{\partial t} = f_i (\mathbf{u}, \frac{\partial \mathbf{u}}{\partial x_1}, \frac{\partial^2 \mathbf{u}}{\partial x_1^2}, ..., \frac{\partial \mathbf{u}}{\partial x_2}, \frac{\partial^2 \mathbf{u}}{\partial x_2^2}, ...); \quad i = 1, ..., d,
\end{equation}
where $f_i$ depends nonlinearly on the state variables and their spatial derivatives.

Our benchmark includes both widely-studied PDE systems in the MD literature and seven systems simulating complex physical processes in fluid dynamics. Previously established datasets include \textit{Advection} and \textit{Burgers} \cite{takamoto2022pdebench}; \textit{Korteweg-de Vries (KdV)}, \textit{Kuramoto-Sivashinsky (KS)},  \textit{Nonlinear Schrödinger (NLS)} , and \textit{Reaction-Diffusion (RD)} \cite{brunton2016discovering}; \textit{Advection-Diffusion (AD)} \cite{both2021deepmod}.

We observe that commonly used PDE systems in the MD literature often fail to capture the complexity of real-world phenomena, which may involve space-dependent or piecewise forcing functions, or exhibit high dimensionality. Therefore, we include the following PDE systems in our testbed: \textit{Heat (Laser)}~\cite{abali2016computational}; \textit{Heat (Solar)}, \textit{Navier-Stokes (Channel)}, \textit{Navier-Stokes (Cylinder)}, and \textit{Reaction-Diffusion (Cylinder)}~\cite{langtangen2017solving}.

A summary of the dataset properties (Table~\ref{tab:pde-summary}) along with full equations and implementation details are included in the supplementary material.

\subsection{Experimental Setup}

\subsubsection{Noise Setup.}

To assess robustness, we simulate measurement noise by corrupting clean state variables with Gaussian noise at signal-to-noise ratios (SNRs) of 40 dB, 30 dB, 20 dB, and 10 dB. We apply multiplicative noise as:
\[
\mathbf{u} \to (1 + \xi) \mathbf{u}, \quad \xi \sim \mathcal{N}(0, \sigma), \quad \sigma = 10^{-\text{SNR}_{\text{dB}} / 20}.
\]
This impacts both the right-hand side and the estimated time derivatives. We compute noisy derivatives using finite differences and evaluate prediction error against the clean derivatives.

\subsubsection{Metrics.}
In order to assess the performance of the discovered equations, we investigate two metrics: 1) \textbf{NMSE} refers to the accuracy of the predicted time derivatives defined as the Normalized Mean Square Error of the derivatives, $\text{NMSE}(\mathbf{u}_t, \mathbf{\hat {\mathbf{u}}}_t) = \frac
{\sum_{i=1}^{N} \left ( u^{(i)}_t - \hat{u}^{(i)}_t \right )^2}
{\sum_{i=1}^{N} \left ( u^{(i)}_t \right )^2  + \epsilon}, 
$
where $\epsilon$ is a small regularization constant set to $10^{-10}$. NMSE measures trajectory-level predictive fidelity. A lower NMSE indicates better agreement between the model and data, and serves as a proxy for how well the discovered dynamics reproduce observed behavior. 2) \textbf{Complexity} refers to the total number of nodes, constant terms, and operations in the expression tree of the equations. This metric encourages parsimony and penalizes unnecessary long or redundant terms. The \texttt{SymPy} package is used to parse the discovered equations and compute their complexities \cite{sympy}.

\subsubsection{Hyperparameter Tuning.}
Since MD methods are sensitive to hyperparameters, we adopt a unified and automated tuning protocol. For each method-dataset pair, we evaluate all combinations from a predefined hyperparameter grid and select the best configuration using a composite fitness score that balances between the complexity and accuracy of the equations. Similar to \cite{merler2024context, shojaee2023transformer}, the fitness function is defined as 
\begin{equation}
    s(f | \mathbf{u}) = \frac{1}{1 + \text{NMSE}(\mathbf{u}_t, f(\mathbf{u}))} + \lambda \exp\left(-\frac{l(f)}{L}\right),
\label{eq:fitness}
\end{equation}
where the hyperparameter $\lambda$ weights the relative importance of accuracy and complexity, $l(.)$ computes the complexity of the equations, and $L$ denotes the maximum equation length. The set of defined hyperparameters for each algorithm is provided in the supplementary materials. Throughout the experiments, we set $\lambda=1$ and $L = 200$.

\subsubsection{Extending Generic Methods to PDEs.} 

Most generic MD methods were originally designed for standard supervised learning tasks, where both input features and target outputs are explicitly available in the datasets. However, in the context of PDE-based dynamical systems, the right-hand side of the equations, which typically involve spatial derivatives, must be constructed from the observed data. To enable the application of these methods to PDE discovery, we preprocess each dataset to compute the required symbolic features.

Specifically, for each PDE system with $d$ state variables, we generate a symbol set that includes the state variables themselves along with their spatial derivatives up to fourth order. These are computed using second-order accurate finite difference approximations via the \texttt{findiff} package~\cite{findiff}. The resulting symbolic feature set is $
\left \{ u_1, u_2, ..., u_d \right \} \cup \left \{ \frac{\partial^j u_i}{\partial x_k^j}\mid i = 1, \ldots, d; j = 1, \ldots, 4; k = 1, \ldots, D \right\}
$. The target variables, i.e., the time derivatives $\partial u_i / \partial t$, are also computed using the same finite difference scheme.

\subsubsection{Training and Inference.}

We split each dataset along the time dimension into three parts: the first 60\% is used for training, the next 20\% for validation (used in hyperparameter tuning), and the final 20\% for testing. After selecting the optimal hyperparameters based on the validation set, each model is retrained on the combined training and validation sets. The final evaluation is performed on the test set using the metrics described earlier.

For GP-based methods, we follow the authors’ recommended default settings and train the models on the combined training and validation data. From the set of candidate expressions generated during evolution, we select the best equation using the fitness function defined in Equation~\ref{eq:fitness}.

To ensure fairness and feasibility, we impose a time limit of 12 hours per experiment, covering both training and hyperparameter tuning. All time-bounded experiments are conducted on an Ubuntu 22.04.3 server equipped with a 24-core Intel Core Ultra 9 285K @ 1.44 GHz CPU and 96 GB of RAM. For methods that require GPU acceleration (EQL, uDSR, ODEFormer, End2end, and DeepMoD), we use a single NVIDIA RTX 4000 Ada Generation GPU with 20 GB of memory.




\subsection{Methods}

From the algorithm categories presented in Section~\ref{sec:related_work}, we selected a representative set of model discovery methods to benchmark. Selection criteria included strong reported performance in prior work and the availability of well-documented, easily adoptable implementations. The studied method include PDEFIND~\cite{rudy2017data}, WSINDy~\cite{messenger2021weak}, ESINDy/EWSINDy~\cite{fasel2022ensemble}, Bayesian~\cite{more2023bayesian}, DeepMoD~\cite{both2021deepmod}, SINDy~\cite{brunton2016discovering}, EQL~\cite{sahoo2018learning}, uDSR~\cite{landajuela2022unified}, PySR~\cite{cranmer2023interpretable}, Operon~\cite{burlacu2020operon}, ODEFormer~\cite{d2024odeformer}, and End2end~\cite{kamienny2022end}. Table~\ref{tab:methods} provides an overview of the selected algorithms, including their methodological type, the systems they are applied to, and a brief description.

Due to memory or runtime limitations, the methods DeepMoD, uDSR, and End2end are unable to handle large datasets efficiently. For these methods, we perform training on a subsampled version of the data, limited to 10,000 points.

\begin{table*}
    \small
    \centering
    \begin{tabular}{lccl}
    \toprule
    \textbf{Method} & \textbf{Type} & \textbf{Systems} & \textbf{Description} \\
    \midrule
        PDEFIND         & LM    & PDE          & Sparse regression on a library of candidate terms for discovering PDEs.       \\
        SINDy            & LM    & ODE          & Sparse regression on a library of candidate terms for discovering ODEs.  \\
        WSINDy           & LM    & PDE          & Weak formulation of PDEFIND for robust identification from noisy or sparse data.         \\
        E(W)SINDy        & LM    & ODE/PDE    & Ensemble-SINDy uses ensembles to quantify the inclusion probability of library terms. \\ 
        Bayesian         & LM    & PDE          & An extension of PDEFIND with variational Bayes to handle noisy data. \\
        DeepMoD          & LM    & PDE          & Joint optimization of function approximation and sparse structure.   \\ \hline
        EQL              & DL    & PDE/ODE      & Shallow neural network with symbolic operators as activation functions. \\
        uDSR             & DL    & PDE/ODE      & Unified GP, LM, and LSPT with a recurrent neural network.  \\ \hline
        PySR             & GP    & PDE/ODE      & Evolutionary algorithm with parallel multi-population support.  \\
        Operon           & GP    & PDE/ODE      & Efficient C++-based GP framework with fine-grained control.  \\ \hline
        ODEFormer        & LSPT  & ODE          & Transformer-based equation discovery from ODE trajectory data. \\
        End2end          & LSPT  & ODE/PDE      & End-to-end transformer model for discovering equations from input-output data. \\
    \bottomrule
    \end{tabular}
    \caption{Overview of model discovery methods benchmarked in MDBench.}
    \label{tab:methods}
\end{table*}

\section{Results and Discussion}
\label{sec:results}

\subsection{Performance on ODE Data}

\begin{figure}
    \centering
    \begin{subfigure}{\linewidth}
        \includegraphics[width=1\linewidth]{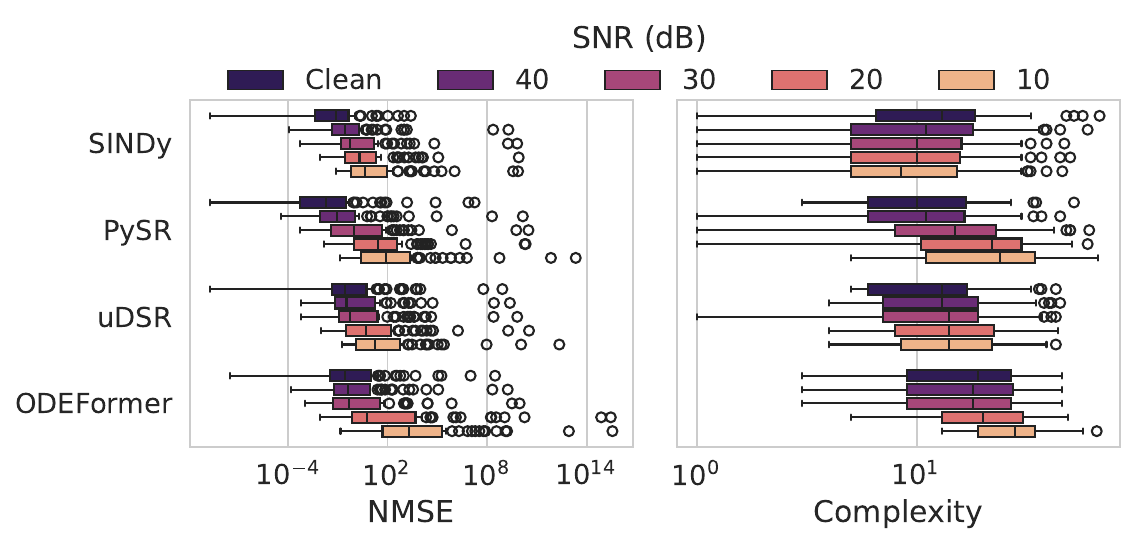}
        \caption{ODE datasets.}
        \label{fig:perf-ode-representative}
    \end{subfigure}
    \begin{subfigure}{\linewidth}
        \includegraphics[width=1\linewidth]{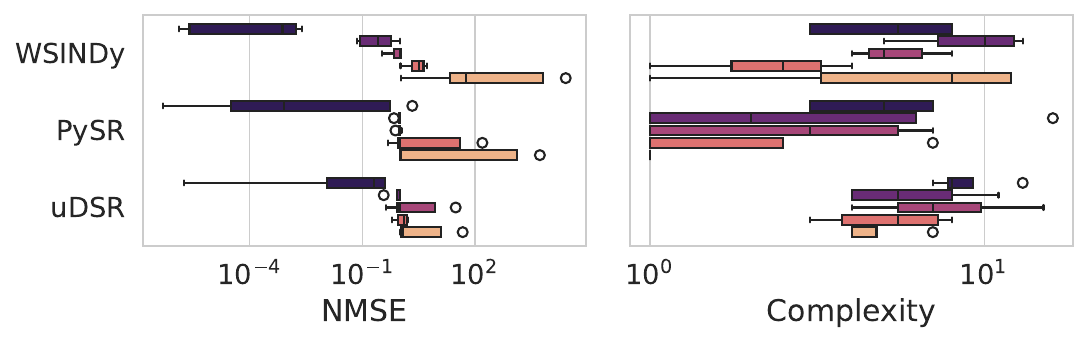}
        \caption{PDE datasets.}
        \label{fig:perf-pde-representative}
    \end{subfigure}
    \caption{Performance box plot for representative MD methods on ODE and PDE datasets.}
    \label{fig:perf-representative}
\end{figure}

Figure \ref{fig:perf-ode-representative} compares the performance of a representative subset of model discovery methods on the ODEBench dataset (see Figure \ref{fig:perf-ode-supp} for additional methods). Most methods were able to recover valid equations under noiseless conditions, but their robustness to increasing noise varied considerably. Tables \ref{tab:ode-performance-1} and \ref{tab:ode-performance-2} in supplementary materials provide a more detailed account of the performance metrics on ODE datasets.

Equations generated by ODEFormer have good predictive accuracy on time derivatives in low-noise settings and clean data (SNR $\geq$ 30 dB), indicating effective generalization. However, its performance degrades rapidly under higher noise levels (SNR $\leq$ 20 dB), suggesting overfitting likely due to its large number of trainable parameters. This is also reflected in the increasing complexity of the discovered equations as noise increases.

In contrast, linear models SINDy and ESINDy generated simpler equations under noisy conditions. The Wilcoxon signed-rank test showed a significant difference between the complexity of SINDy equations and that of EQL (next simplest equations) for noisy data $W=6951.5, p<10^{-8}$. The included sparsity-promoting regularization acts as an implicit denoising method, suppressing uncertain or spurious terms introduced by noise. This behavior, while leading to less complex equations, may also result in the omission of relevant nonlinear dynamics, especially at low SNR.

GP-based methods generally achieved lower error in predicted time derivatives compared to other methods in low-noise settings and clean data (SNR $\geq$ 30 dB). For instance, PySR has a significantly lower NMSE than ESINDy based on Wilcoxon signed-rank test (log-scaled, $W=9774, p \leq 0.0001$). GP algorithms showed a tendency to overfit in noisy settings, producing highly complex and inaccurate expressions. This might suggest that GP-based methods lacking explicit regularization or those imposing minimal assumptions on the equation skeleton in the search space are more sensitive to noisy inputs.

End2end exhibited the highest predictive errors among the evaluated methods, indicating that it struggled to capture the underlying system dynamics (Wilcoxon signed-rank test with EQL, log scale NMSE, $W=12795, p<10^{-10}$). This may be due to the fact that its pretrained model was not exposed to any dynamical systems during training, limiting its ability to generalize to such tasks.

As system dimensionality increases, most methods exhibit a corresponding rise in training time. Among the evaluated approaches, SINDy and Operon are the most computationally efficient, whereas EQL and uDSR are significantly slower, differing by several orders of magnitude. Figure \ref{fig:ode-train-time} in supplementary materials shows the training times of various MD methods on ODE datasets. 


\subsection{Performance on PDE Data}

Compared to ODEs, PDE systems present a more challenging testbed due to the presence of spatial derivatives and higher-order terms, which tend to amplify noise. Figure \ref{fig:perf-pde-representative} shows the aggregate NMSE and complexity of representative methods over a selected subset of PDE datasets for which all methods completed successfully (refer to Figure \ref{fig:perf-pde-supp} for all methods). The selected datasets include \textit{Advection}, \textit{Burgers}, \textit{KdV}, \textit{KS}, \textit{Heat (Solar) 1D}, and \textit{AD}. The primary reasons for a method's failure to discover an underlying model include: a lack of support for higher-dimensional data in the published software, failing to finish training before the timeout threshold was reached, and runtime errors within the published codebases. EQL and End2end are excluded from the PDE evaluations due to repeated failures, mainly caused by implementation issues or limitations in handling high-dimensional inputs. In particular, the pretrained transformer model for the End2end method has been trained on synthetic equations with up to 10 features \cite{kamienny2022end}, which prohibits its usage on high-dimensional systems or systems with more than two state variables.

Figure \ref{fig:perf-pde-representative} shows that there is a sharp error gap between the NMSE for clean and low-noise systems (SNR = 40 dB) among the selected PDE datasets. The logarithm of the error ratio between the noisy and clean settings for PDEs is $3.13 \pm 2.09$, compared to $0.88 \pm 2.12$ for the ODE datasets. This illustrates that the accuracy of the predicted time derivatives estimated by model discovery methods for PDE systems is more sensitive to noise than for ODE systems. One reason is that while ODE systems involve only a single time derivative per state variable, PDE systems include higher-order and spatial derivatives. These high-order derivatives amplify noise in measured data, leading to greater accumulation of the errors.

Most LM methods, except for DeepMoD, achieved lower NMSE and better robustness in low-noise settings (SNR $\geq$ 30 dB) and noise-free data compared to GP methods. For instance, WSINDy obtained significantly lower error compared to PySR ($W=455.0, p < 10^{-4}$) and Operon ($W=466.0, p < 10^{-4}$) in the mentioned noise settings. PySR and Operon, while producing competitive results in clean settings, suffer from reduced predictive fidelity as noise increases. Despite simpler equations, uDSR tends to generate the most inaccurate equations in the noise-free setting.

Table \ref{tab:pde-performance} summarizes the NMSE and expression complexity for each method across all PDE datasets in the clean (noise-free) setting. Almost all methods achieve low prediction error on the \textit{Heat (Solar) 1D} dataset. However, as the spatial dimension of the \textit{Heat (Solar)} system increases (from 1D to 3D), the NMSE of predicted derivatives decreases. This is because higher dimensions roughly double the size of the function library for sparse-regression methods and the number of symbols for GP-based methods, thereby enlarging the search space and making model discovery more challenging. These results highlight the need for developing scalable algorithms capable of handling high-dimensional data.

\begin{table*}[ht]
\small
\centering
\renewcommand{\arraystretch}{1.2}
\begin{tabular}{l|cccccccc}
\toprule
\textbf{Dataset} & PDEFIND & Bayesian & WSINDy & EWSINDy & DeepMoD & PySR & Operon & uDSR \\
\midrule
Advection & \checkmark & $10^{-6} (21)$ & \checkmark & \checkmark & $10^1 (12)$ & \checkmark & \checkmark & $10^{-6} (8)$\\
Burgers & \checkmark & \underline{$10^{-6} (9)$} & \checkmark & \checkmark & \checkmark & \checkmark & \checkmark & $10^{-1} (8)$\\
KdV & \checkmark & \underline{$10^{-3} (9)$} & \checkmark & \checkmark & \checkmark & \checkmark & $10^{-1} (7)$ & $10^{-1} (7)$\\
KS & \checkmark & \underline{$10^{-3} (12)$} & \checkmark & \checkmark & $10^1 (3)$ & \checkmark & \underline{$10^{-3} (23)$} & $10^{0} (4)$\\
AD &  \underline{$10^{-5} (17)$} & \underline{$10^{-5} (12)$} & \checkmark & \checkmark & $10^1 (3)$ & \checkmark & \checkmark & $10^{0} (15)$\\
Heat (Solar) 1D & \checkmark & $10^{-3} (8)$ & \checkmark & \checkmark  & $10^6 (3)$ & \checkmark & \checkmark & $10^{-2} (13)$\\
Heat (Solar) 2D & $10^{-2} (30)$ & - & $10^{-3} (17)$ & $10^{-3} (9)$ & - & \underline{$10^{-3} (3)$} & $10^{-3} (7)$ & $10^{-2} (12)$\\
Heat (Solar) 3D & $10^1 (23)$ &  - & - & - & - & $10^{0} (10)$ & $10^{0} (15)$ & $10^{0} (3)$\\
Heat (Laser) & $10^0 (14)$ & - & - & - & - & $10^{0} (11)$ & $10^{0} (15)$ & $10^{0} (6)$ \\
NLS & $10^{-1} (16)$ & - & $10^{-1} (16)$ & - & - & $10^{0} (29)$ & $10^{-1} (53)$ & $10^{2} (22)$\\
RD & $10^{-3} (6) $ & - & $10^{-3} (6)$ & - & - & $10^{-2} (5)$ & $10^{-2} (14)$ & $10^{-2} (12)$ \\
NS (Channel) & $10^{2} (82)$ & - & - & - & - & $10^{-3} (4)$ & $10^{-3} (15)$ & $10^{-3} (22)$ \\
NS (Cylinder) & $10^{-2} (42)$ & - & $10^{-2} (95)$ & - & - & $10^{-1} (25)$ & $10^{-1} (30)$ & $10^{-1} (38)$ \\
RD (Cylinder) & - & - & $10^{-2} (3253)$ & $10^{-1} (1603) $& - & $10^{-1} (27)$ & $10^{-2} (51)$ & $10^{-1} (50)$\\
\bottomrule
Runtime (min) & $26 \pm 53$ & $102 \pm 205$ & $4 \pm 6$ & $94 \pm 87$ & $108 \pm 131$ & $8 \pm 13$ & $54 \pm 104$ & $9 \pm 11$ \\
\bottomrule
\end{tabular}
\caption{Performance of MD methods on PDE datasets, measured by the NMSE of the time derivatives on the test set. Equation complexity is given in parenthesis. Reported NMSEs are rounded to the nearest power of 10. A check mark (\checkmark) indicates successful identification of the full equation, with all coefficients being in $\pm5\%$ of ground truth. Underlined entries denote partially correct equations, where only one term is missing or extra, and the remaining terms have coefficients that are within $\pm5\%$ of ground truth. Empty entries (-) show method failure or timeout. The table includes the average and standard deviation of method runtimes on clean data (in minutes).}
\label{tab:pde-performance}
\end{table*}

The datasets \textit{Heat (Laser)} and \textit{RD (Cylinder)} are more challenging than the rest of the PDEs in the benchmark due to their use of custom forcing functions. The \textit{Heat (Laser)} system contains a spatially-dependent forcing function, while the \textit{RD (Cylinder)} system not only includes six state variables but also has piecewise linear forcing functions that vary across the spatial domain. Most of the LM methods fail at generating equations for these datasets likely because of the huge library size and sparse regression problem. The results of PySR, Operon, and uDSR methods on these datasets illustrate that existing model discovery methods perform poorly on such datasets, suggesting the need for more generalizable and robust algorithms for discovering real-world PDE systems.


Runtimes vary widely by method and PDE dataset, ranging from a few seconds to several hours (Table \ref{tab:pde-performance}). Overall, PDEFind, WSINDy, PySR, and uDSR demonstrate the fastest runtimes. While GP methods, particularly PySR, tend to be slower than linear methods on 1D problems, they scale more favorably in higher dimensions. For example, on the \textit{Heat (Solar)} dataset, PDEFind is approximately $40\times$ and $465 \times$ slower in the 2D and 3D cases, respectively, compared to the 1D case. In contrast, PySR slows down by about $33\times$ in 2D and only $12\times$ in 3D. This disparity likely stems from the growing size of sparse regression systems encountered by linear model-based methods. This suggests that in general, in time-constrained environments, LM methods are a better choice for low-dimensional datasets, while GP methods and uDSR scale more robustly to higher-dimensional data. More granular runtime data can be found in Table \ref{tab:pde-times}.

\subsection{Limitations}
\label{sec:limitations}

Despite recent progress in data-driven model discovery, we highlight several key limitations.

First, many algorithms implicitly assume uniform physical parameters (e.g., diffusivity, conductivity) across space and time. This assumption simplifies inference but fails to reflect heterogeneities in real-world systems such as \textit{Heat (Laser)} and \textit{Reaction-Diffusion (Cylinder)} systems (Table \ref{tab:pde-performance}). Models built under this assumption may generalize poorly to such heterogeneous environments.

Second, current methods show degraded performance on systems with many state variables or high-dimensional spaces. As shown in Table \ref{tab:pde-performance}, systems with one state variable have a much higher successful discovery rate and lower errors on average compared to systems with more than one state variable. In the case of \textit{Heat (Solar)} systems, the error increases for all methods as the spatial dimension grows. LM approaches suffer from the exponential growth of function libraries, while GP-based methods face intractable search spaces. This scalability bottleneck limits applicability to realistic systems such as multi-component dynamical systems or high-resolution simulations.

Third, most discovery methods operate on numerical approximations of derivatives, making them highly sensitive to noise. In PDE systems, the amplification of noise through higher-order derivatives further compromises model fidelity as can be seen from Figure \ref{fig:perf-pde-supp} in supplementary materials. Without robust denoising mechanisms or noise-aware formulations, the discovered equations may be structurally incorrect despite yielding a low NMSE on predicted time derivatives.

Fourth, there is no established metric for quantifying equation fidelity. Standard metrics such as NMSE and symbolic complexity do not always reflect the symbolic correctness of discovered equations. For example, while GP methods achieved a relatively low NMSE on \textit{NS (Channel)} dataset, the discovered equations do not capture the true dynamics. Without a robust metric that captures functional equivalence, e.g., using a discretized version of the Sobolev semi-norm for PDE systems \cite{schaback2015computational}, benchmarking results may obscure deeper algorithmic failures.

Finally, several methods fail on specific datasets due to implementation limitations (e.g., lack of support for multi-dimensional systems, unstable solvers, or uninformative error messages). This undermines reproducibility and suggests the need for compatible implementations that can be used by other researchers to apply the methods on their datasets.

\section{Conclusion}
\label{sec:conclusion}
In this paper, we introduced MDBench, a comprehensive and extensible benchmark for evaluating 12 model discovery methods across 77 dynamical systems. 
We evaluated the quality of the discovered dynamics across four methodological classes, assessing predictive accuracy of time derivatives, equation complexity, and fidelity to ground-truth dynamics. 
We release MDBench as an open-source, extensible benchmarking framework as well as a representative collection of datasets, and believe it provides a foundation for the evaluation and development of novel model discovery algorithms for dynamical systems. We hope MDBench serves as a foundation for future research in interpretable, robust model discovery across complex dynamical systems.

\clearpage

\appendix
\setcounter{figure}{0}
\renewcommand{\thefigure}{A\arabic{figure}}
\setcounter{table}{0}
\renewcommand{\thetable}{A\arabic{table}}


\section{MDBench Pipeline}

We release the MDBench pipeline as an open-source tool to support reproducible, extensible research in model discovery. The framework is modular by design — drawing inspiration from SRBench~\cite{la2021contemporary} — and is structured to simplify the integration of new algorithms and datasets. All datasets are provided in a unified format to ensure cross-method compatibility and ease of adoption by the broader research community.

To promote reproducibility, we provide isolated environments for each method, with exact package versions and dependencies specified. The datasets and the complete pipeline, including environment setup, algorithm installation, noise configuration, and experiment execution, are publicly available.

The MDBench repository is organized to easily support integration of new methods. Implementations are located in \texttt{mdbench/algorithms/}. Methods under \texttt{sr} are general-purpose model discovery algorithms applicable to both ODE and PDE systems. The \texttt{pde} and \texttt{ode} directories contain methods specific to PDE and ODE datasets, respectively. See the main \texttt{README.md} for guidance on adding new methods to the benchmark.

The command \texttt{./install.sh} creates a separate Conda environment for each algorithm and installs all required dependencies. The script \texttt{run.sh} serves as the main entry point for training. For example,
\texttt{./run.sh --algorithm operon --data\_type ode --dataset\_path data/ --n\_jobs 20}
launches 20 parallel processes, trains the Operon method on all ODE datasets in the \texttt{data} directory, and stores results in \texttt{[method\_name]-[data\_type].jsonl} files. Use \texttt{./run.sh -h} to see all available options.

Datasets are stored in compressed \texttt{NumPy} format (\texttt{.npz}). Filenames indicate the dataset name and the SNR level for noisy variants. Clean derivatives, computed from the ground-truth differential equations for ODEs, and via finite differences for PDEs, are included to enable evaluation of predictive accuracy.


\section{Datasets}
Table \ref{tab:pde-summary} provides an overview of PDE systems in the benchmark. The newly generated datasets require the legacy FEniCs package \cite{fenics}, which can be readily installed on Ubuntu-based systems. Please refer to the README in the \texttt{/scripts} subfolder of the MDBench codebase for additional guidance on using FEniCS to generate the datasets, as well as the dataset-specific instructions below.

\begin{table*}
\centering
\renewcommand{\arraystretch}{1.1}
\begin{tabular}{lccccc}
\hline
\textbf{Dataset} & \textbf{$N_d$} & \textbf{$N_t$} & $\Delta t$ & \textbf{$N_s$} & $\Omega$  \\
\hline
Advection & 1 & 201 & 0.01 & 1024 & $x \in [0.00049, 0.9995]$ \\
Burgers & 1 & 101 & 0.1 & 256 & $x \in [-8, 8)$ \\
KdV & 1 & 201 & 0.1 & 512 & $x \in [-30, 30)$ \\
KS & 1 & 251 & 0.4 & 1024 & $x \in [0.098, 100.53]$ \\
AD & 1 & 61 & 0.1 & 51 $\times$ 51 & $x \in [-5, 5], y \in [-5, 5]$\\
Heat (Solar) 1D & 1 & 576 & 300 & 51  & $x \in [-1.5, 0] $\\
Heat (Solar) 2D & 1 & 576 & 300 & 51 $\times$ 51 & $x \in [-0.375, 0.375], y \in [-1.5, 0]$ \\
Heat (Solar) 3D & 1 & 20 & 300 &  51 $\times$ 51 $\times$ 11 & $x \in [-0.375, 0.375], y \in [-0.375, 0.375], z \in [-1.5, 0] $ \\
Heat (Laser) & 1 & 20 & 0.1 & 201 $\times$ 201 $\times$ 3 & $x \in [0, 0.1], y \in [0, 0.1], z \in [0, 0.001] $ \\
NLS & 2 & 251 & 0.0126 & 256 & $x \in [-5, 5)$ \\
RD & 2 & 100 & 0.101 & 32 $\times$ 32 & $x \in [-10, 10), y \in [-10, 10)$\\
NS (Channel) & 2 & 50 & 0.02 & 9 $\times$ 9 & $x \in [0.25, 0.75], y \in [0.25, 0.75]$ \\
NS (Cylinder) & 3 & 51 & 0.02 & 100 $\times$ 30 & $x \in [0.441, 1.316], y \in [0.084, 0.326]$\\
RD (Cylinder) & 6 & 51 & 0.02 & 100 $\times$ 30 & $x \in [0.441, 1.316], y \in [0.084, 0.326]$\\
\hline
\end{tabular}
\caption{
An overview of PDE datasets properties, where $N_d$ denotes number of state variables, $N_t$ represents number of time points, $\Delta t$ shows time steps, $N_s$ is spatial resolution, and $\Omega$ represents the spatial domain. Note that spacing between time points and spatial grid along each coordinate is uniformly spaced.}
\label{tab:pde-summary}
\end{table*}

\subsection{Advection}
The advection system, adopted from \cite{takamoto2022pdebench}, is a simple linear system in 1D space with the equation
\begin{align*}
    \frac{\partial u(t, x)}{\partial t} + \beta \frac{\partial (t, x)}{\partial x} = 0,
\end{align*}
where $\beta = 0.1$ denotes constant advection speed. The boundary condition is periodic and the initial condition is a superposition of sinusoidal waves
\begin{align*}
    u(0, x) = \sum_{k_i = k_1, k_2} A_i \sin (k_i x + \phi_i),
\end{align*}
where $k_i = 2 \pi n_i$ are wave numbers with $n_i$ being random integer numbers between [1, 8], and $A_i$ and $\phi_i$ are floating point numbers uniformly chosen between [0, 1] and $(0, 2\phi)$, respectively.

\subsection{Burgers}
The Burgers equation, from \cite{takamoto2022pdebench}, models the diffusion process in fluid dynamics as
\begin{gather*}
    \frac{\partial u(t, x)}{\partial t} + u(t, x)\frac{\partial (t, x)}{\partial x} - \nu \frac{\partial^2 u(t, x)}{\partial x^2} = 0,
\end{gather*}
where $\nu = 0.1$ is the constant diffusion coefficient.

The initial and boundary conditions are similar to the Advection system.

\subsection{Korteweg-de Vries (KdV)}
The KdV system, taken from \cite{brunton2016discovering}, models unidirectional propagation of small-amplitude waves in shallow water. The governing equation is given by
\begin{gather*}
\frac{\partial u(t, x)}{\partial t} + 6 u(t, x)\frac{\partial u(t, x)}{\partial x} + \frac{\partial^3 u(t, x)}{\partial x^3} = 0,
\end{gather*}

where the initial condition is constructed from two KdV solutions of different amplitudes in order to distinguish the solution from the standard advection equation. The initial condition is a superposition of squared hyperbolic secant equations
\begin{gather*}
u(0, x) = \sum_{c_i = 1, 5}\frac{c_i}{2} \text{sech}^2 \left ( \frac{\sqrt{2}}{2} \left (x - x_0\right )\right ),
\end{gather*}
where $c_i$ are the speeds at which the waves travel.

\subsection{Kuramoto-Sivashinsky (KS)}

The KS equation is known for modeling the dynamics of various physical systems such as fluctuations in fluid films on inclines \cite{brunton2016discovering}. This equation includes a fourth-order diffusion term as a diffusive regularization of non-linear dynamics. The system is governed by the equation

\begin{gather*}
    \frac{\partial u(t, x)}{\partial t} = -u(t, x) \frac{\partial u(t, x)}{\partial x} - \frac{\partial^2 u(t, x)}{\partial x^2} -\frac{\partial^4 u(t, x)}{\partial x^4}.
\end{gather*}
\subsection{Nonlinear Schrödinger (NLS)}
The NLS system appears in optics such as studying nonlinear wave propagation optical fibers. The complex-valued equation follows
\[
    \frac{\partial u(t, x)}{\partial t} = 0.5 i \frac{\partial^2 (t, x)}{\partial^2 x} + i |u(t, x)|^2 u(t, x).
\]
Since many of the model discovery implementations do not support complex numbers, the complex-valued system is decomposed into a real-valued, coupled PDE system

\begin{align*}
    \frac{\partial a(t, x)}{\partial t} &= -0.5\frac{\partial^2 b(t, x)}{\partial x^2} - a^2(t, x)b(t, x) - b^3(t, x), \\
    \frac{\partial b(t, x)}{\partial t} &=  0.5\frac{\partial^2 a(t, x)}{\partial x^2} + a(t, x)b^2(t, x) + a^3(t, x).
\end{align*}

\subsection{Reaction-Diffusion}
Reaction diffusion systems are used extensively to study pattern forming systems in physics \cite{brunton2016discovering}. One particular class of DR systems with nonlinear coupling terms are governed by the coupled PDE

\begin{align*}
\frac{\partial u}{\partial t} &= 0.1\nabla^2 u + \lambda(A)u - \omega(A)v \\
\frac{\partial v}{\partial t} &= 0.1\nabla^2 v + \omega(A)u + \lambda(A)v \\
A &= u^2 + v^2, \quad \omega = -\beta A^2, \quad \lambda = 1 - A^2,
\end{align*}
where $\beta = 1$ is a coefficient that controls the strength of the nonlinear coupling between the $u$ and $v$ fields.

\subsection{Advection-Diffusion}

The Advection-Diffusion system, adopted from \cite{both2021deepmod}, appears in heat transfer, pollution transport, and chemical physics, to name a few. We study the 2D Advection-Diffusion system

\begin{align*}
    \frac{\partial u}{\partial t} = - \nabla \cdot (-D \nabla u + v. u),
\end{align*}
where $v = (0.25, 0.5)$ is the constant velocity vector describing the advection, and $D = 0.5$ denotes the diffusion coefficient.

\subsection{Heat (Solar)}
The \textit{Heat (Solar)} dataset from \cite{logg2012automated} simulates the heating of the earth's surface by solar radiation.  For a given dimension $d\in\{1,2,3\}$, a rectangular domain $\omega$ is defined with $x_{d-1}=0$ representing the earth's surface. Using default settings, the domain will be 1.5m in depth and 0.75m in width (for 2D and 3D problems). In the 3D case, we sample the first 20 time points. 

Formally, the \textit{Heat (Solar)} system is the solution to the following initial-boundary value problem:
\begin{align*}
\rho c \frac{\partial u}{\partial t} &= \kappa \nabla^2u\\
u(x_0,\ldots,0)&=T(t)\\
\frac{\partial u}{\partial x_i} &= 0 \;\mathrm{for}\; i\neq d-1\\
T&=u_0
\end{align*}
The function $T(t)=T_R+T_A\sin(\omega t)$ represents surface temperature boundary condition. The parameters of the model were chosen to be representative of the earth's surface:
\begin{itemize}
    \item $T_R=10\;^\circ$C, the reference temperature;
    \item $T_A=10\;^\circ$C, the amplitude of temperature variation;
    \item $\omega=7.27\times 10^{-5}$ Hz, the frequency of temperature variation, $\approx 2\pi/86,400$, the number of seconds in one day;
    \item $\rho=1500$ kg/m$^3$, soil density; 
    \item $c=1600$ N$\cdot $m/kg, heat capacity; and
    \item $\kappa=2.3$ N/K$\cdot$s, thermal conductivity. 
\end{itemize}
The model captures the temperature variations in the soil at five minute intervals over two days, resulting in 576 timesteps. The mesh is uniformly spaced with 50 subdivisions in 1D, of size $50\times 50$ grid in 2D, and of size $50\times 50\times 10$ in 3D.  The dataset can be reproduced by running the script \texttt{fenics\_heat\_soil\_uniform.py}.
\begin{figure}[h!]
    \centering
    \includegraphics[width=.5\linewidth]{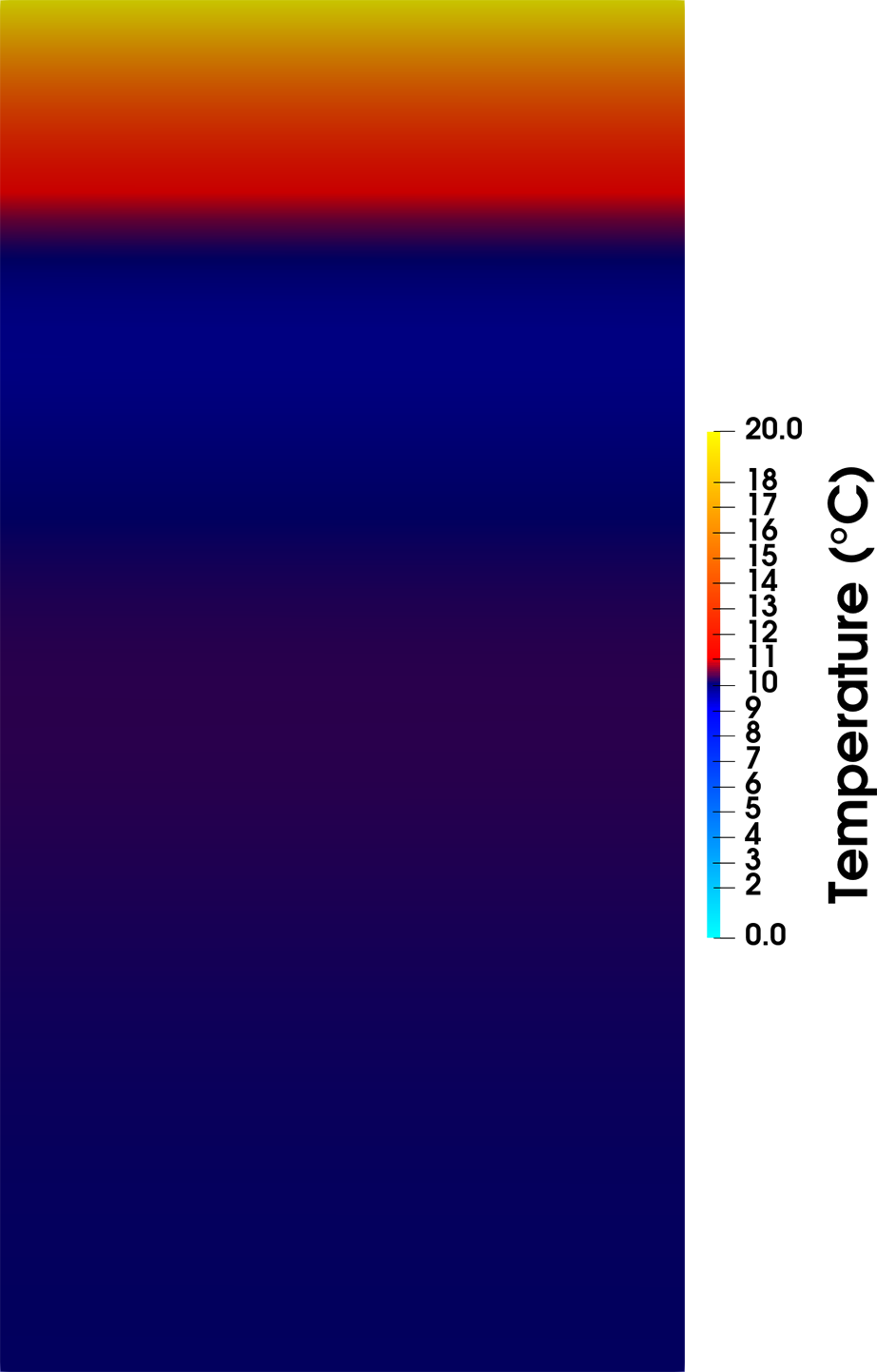}
    \caption{The state of the \textit{Heat (Solar) 2D} system at $t=360$.}
    \label{fig:heatsolar2d}
\end{figure}
\subsection{Heat (Laser)}
The \textit{Heat (Laser)} dataset from \cite{abali2016computational} simulates the dynamics of heat transfer through a rigid body with a custom heat source. In this dataset, the heat supply comes from a moving laser beam that is used to heat a 3D plate in a concentrated manner. The power of the laser follows
\[
L(t, x, y) = P e^{-k \left ( 
\left ( x - \frac{l}{2}(1 + \frac{1}{2} \sin(2 \pi \tau ) \right )^2 +
(y - v_L t)^2
\right )}, 
\]
where $P = 3,000\; \text{W/Kg}$ represents the power of the laser beam and $k=5 \times 10^4$ denotes a large coefficient used to make the beam effect local. The laser moves sinusoidally along $x$ with a time parameter $\tau = t/t_{\text{end}}$ and linearly along $y$ with a constant speed $v_L = 0.02\;\text{m/s}$.

The temperature dynamics follows the equation
\[
\rho c \frac{\partial T}{\partial t} - \kappa \nabla^2 T - \rho L = 0,
\]
where $T(t, x, t)$ is the temperature of the plate, $c = 624 \text{J/kg}$ denotes heat capacity, $\kappa = 30.1 \text{W/m}$ refers to thermal conductivity, and $\rho = 7860\text{kg/m}^3$ is mass density of steel.

The dataset can be reproduced from the script \texttt{fenics\_heat3d\_laser.py}. The original dataset simulates the dynamics for $t \in [0, 50], \Delta t  = 0.1$. We sample the first 20 time points in our analysis. 
\begin{figure}[h!]
    \centering
    \includegraphics[width=\linewidth]{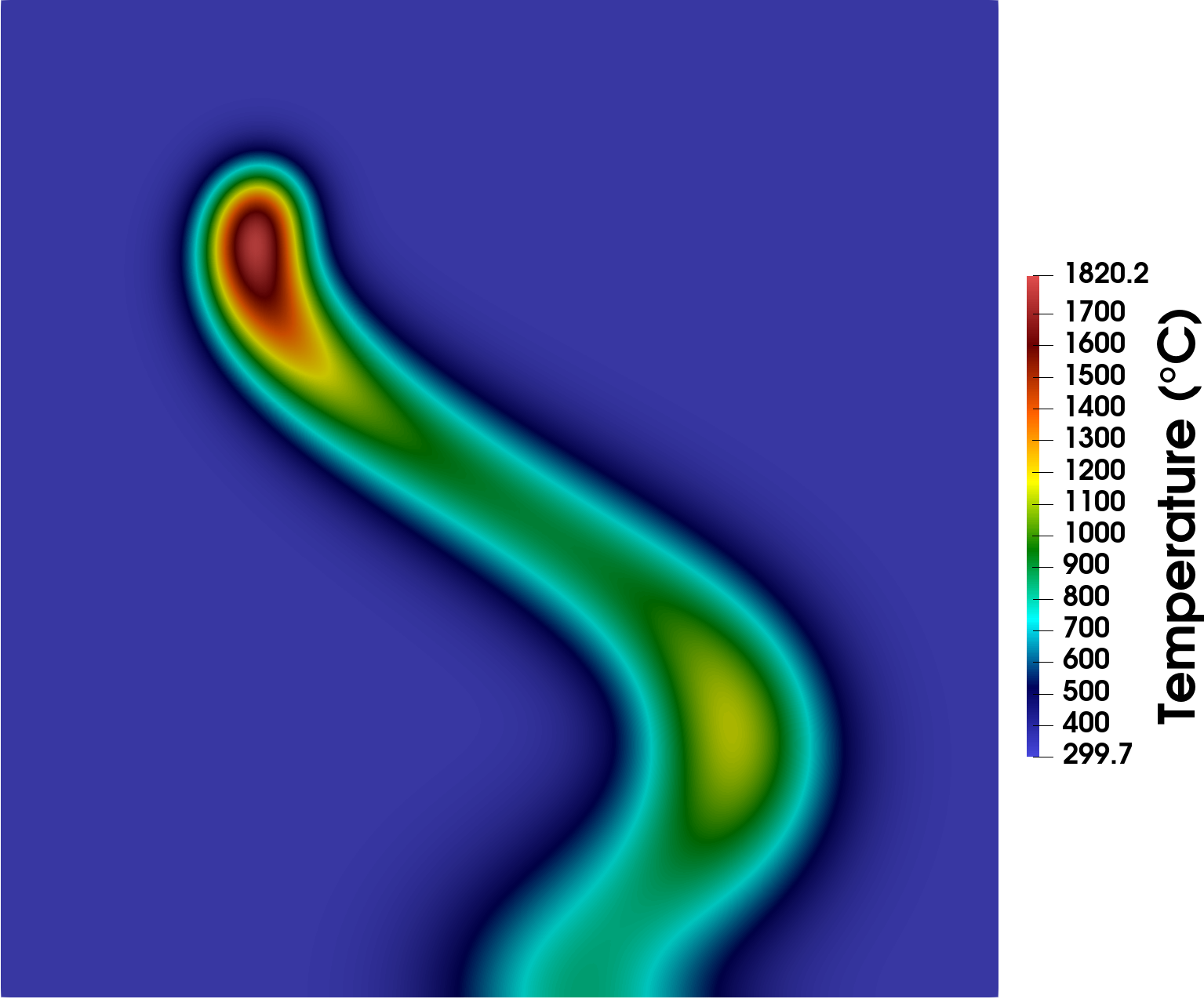}
    \caption{The state of the \textit{Heat (Laser)} system at $t=40$.}
    \label{fig:heatlaser}
\end{figure}
\subsection{Navier-Stokes Channel (NS Channel)}
The incompressible Navier-Stokes equations model the motion of viscous fluids. The \textit{NS Channel} dataset, adopted from \cite{langtangen2017solving}, simulates a liquid traveling through a 2D channel along the $x$ direction. The dynamics follow the system of equations
\begin{align*}
\rho \left( \frac{\partial \mathbf{u}}{\partial t} + (\mathbf{u} \cdot \nabla)\mathbf{u} \right) &= \nabla \cdot \sigma(\mathbf{u}, p) \\
\nabla \cdot \mathbf{u} & = 0,
\end{align*}
where $\sigma (\mathbf{u}, p) = \mu \left ( \nabla \mathbf{u} + \left ( \nabla \mathbf{u} \right )^T  \right ) - p \mathbf{I} $ denotes the stress tensor for a Newtonian fluid, $\mathbf{u}(x,y)$ represents the velocity field, $p(x,y)$ is the pressure field, $\rho = 1$ is density, and $\mu = 1$ is kinematic viscosity. The boundary conditions are: $\mathbf{u}(x,0)=\mathbf{u}(x,1)=0$, i.e., no velocity at the walls of the channel; $p(0,y)=8$ at the channel inflow, and, $p(1,y)=0$ at the channel outflow.


The full dataset is simulated for $t \in [0, 10], \Delta t = 0.02$, with $16\times16$ uniform mesh defined over the unit square channel. The dataset can be reproduced from the script \texttt{fenics\_navier\_stokes\_channel.py}. In our benchmark, we subsample a $9 \times 9$ spatial grid centered around the original grid from the 10th time point to the 60th time point.

\begin{figure}[h!]
    \centering
    \includegraphics[width=\linewidth]{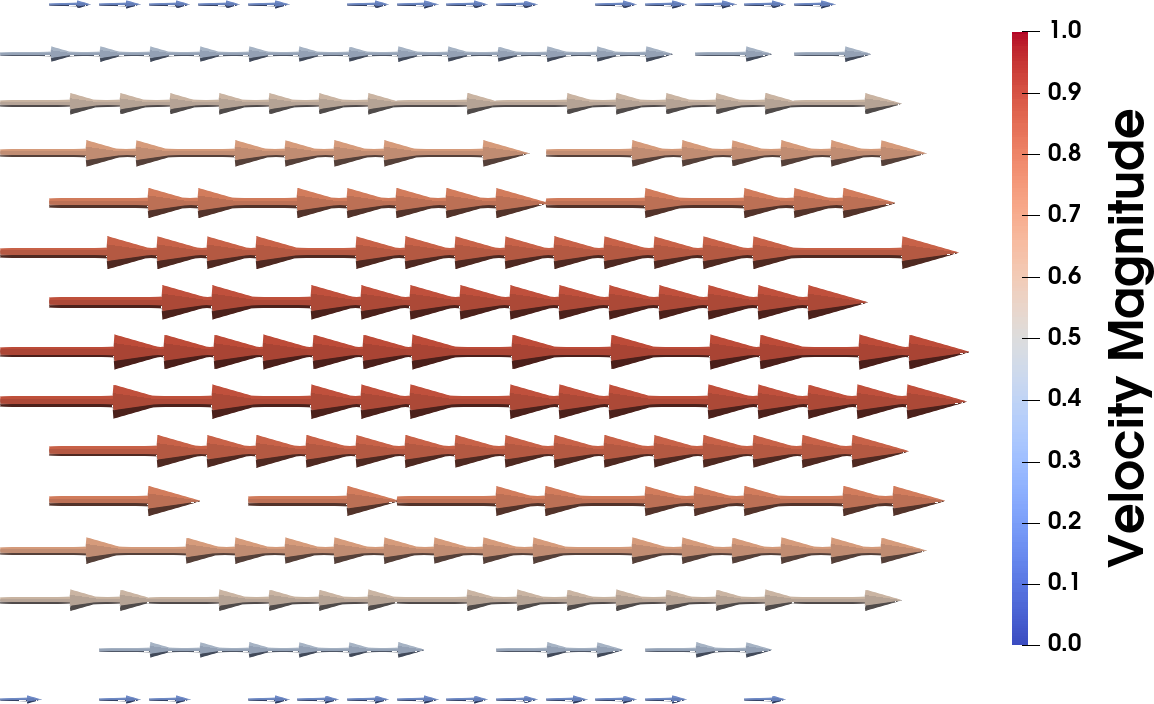}
    \caption{The state of the \textit{NS Channel} system at $t=10$.}
    \label{fig:nschannel}
\end{figure}

\subsection{Navier-Stokes Cylinder (NS Cylinder)}
The \textit{NS Cylinder} dataset is another example of a Navier-Stokes system as published in the FEATFLOW Benchmark Suite \cite{featflow}. The system records the velocity and pressure generated by flow through a 2D rectangular channel 2.2 m in length and 0.41 m in width, with a 5 cm wide cylinder obstruction centered at $xy$-coordinates $(0.2,0.2)$. In this system, the fluid density $\rho = 1$ and the kinematic velocity $\mu = 0.001$. With a maximum velocity of 1.5 m/s, the system has a Reynolds number $Re=100$, which results in turbulent flow, with vortex shedding beyond the cylinder.

\begin{figure}[h!]
    \centering
    \includegraphics[width=\linewidth]{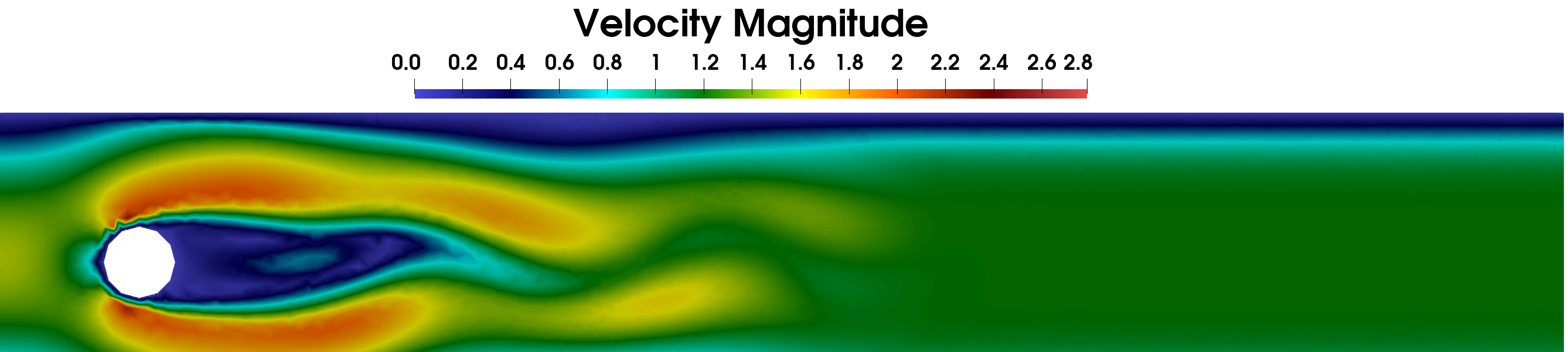}
    \caption{Velocity of the \textit{NS Cylinder} system at $t=999$.}
    \label{fig:nscylindera}
\end{figure}   
\begin{figure}[h!]
    \centering
    \includegraphics[width=\linewidth]{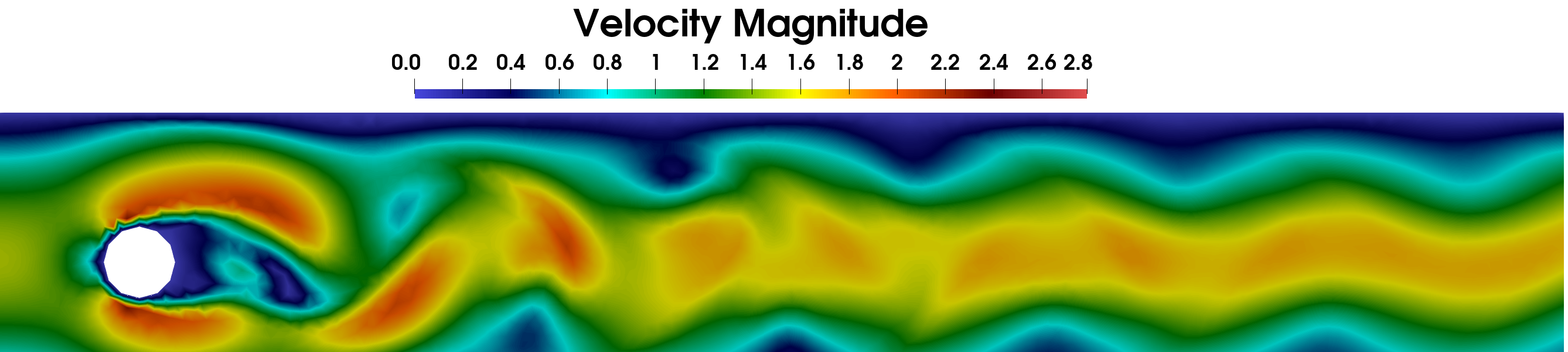}
    \caption{Velocity of the \textit{NS Cylinder} system at $t=4,999$.}
    \label{fig:nscylinderb}
\end{figure}

\begin{figure}
    \centering
    \includegraphics[width=\linewidth]{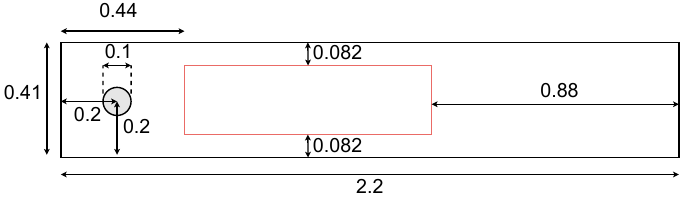}
    \caption{Geometry for the \textit{NS Cylinder} and \textit{RD Cylinder} systems along with the subsampling grid.}
    \label{fig:cylinder-grid}
\end{figure}

The unstructured mesh for the \textit{NS Cylinder} dataset was generated using the \texttt{mshr} subpackage of FEniCS \cite{fenics}. The dataset simulates the system over 5 seconds, saving the data at 1,000 $Hz$ for a total of 5,000 timesteps. The dataset can be reproduced from the script \texttt{fenics\_navier\_stokes\_cylinder.py}. Please note that the dataset must be converted into a structured mesh on a uniform grid prior to use for model discovery. The structured data can be generated from the unstructured version using the \texttt{ns\_cylinder\_structured.py} script. In our analysis, we subsample a rectangular grid in front of the cylinder (Figure \ref{fig:cylinder-grid}) in space and $t \in [1, 2], \Delta t = 0.02$ in time.

\subsection{Reaction-Diffusion (Cylinder)}
\begin{figure}[h!]
    \centering
    \includegraphics[width=\linewidth]{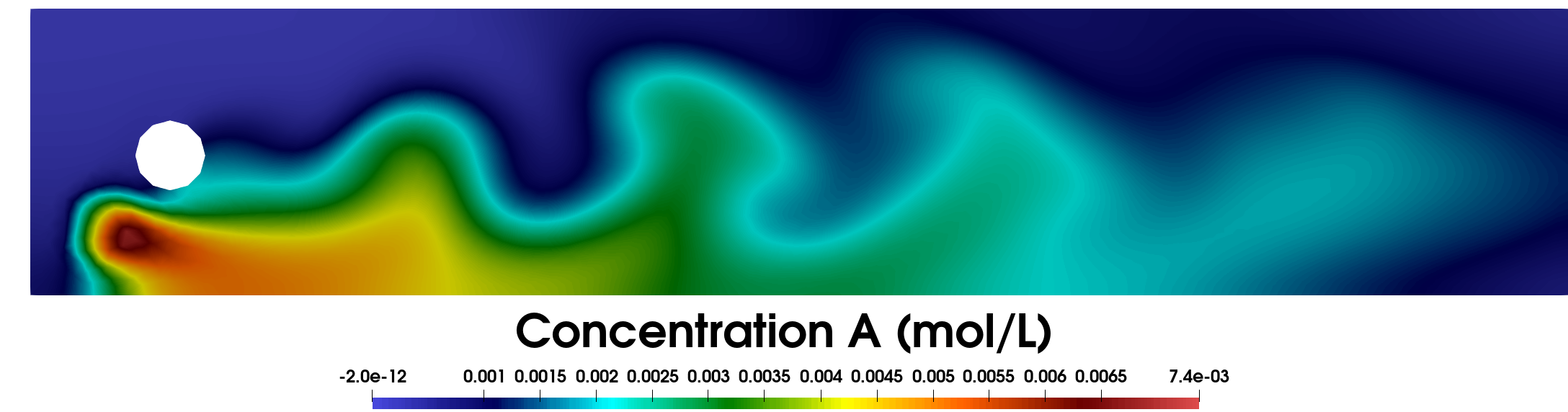}
    \caption{Concentration of \textit{A} in the \textit{Reaction-Diffusion Cylinder} system at $t=199$.}
    \label{fig:diffusionreactiona}
\end{figure}

\begin{figure}[h!]
    \centering
    \includegraphics[width=\linewidth]{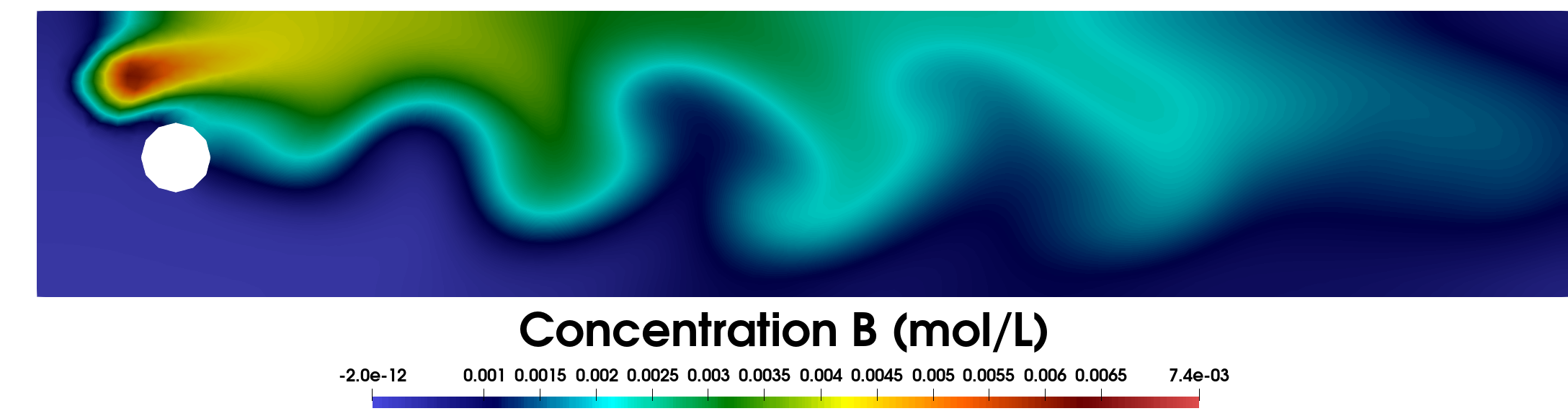}
    \caption{Concentration of \textit{B} in the \textit{Reaction-Diffusion Cylinder} system at $t=199$.}
    \label{fig:diffusionreactionb}
\end{figure}

The \textit{Reaction-Diffusion Cylinder} dataset utilizes the velocity component $\mathbf{v}$ of the solution to the \textit{NS Cylinder} system to simulation an advection-diffusion-reaction system as detailed in \cite{langtangen2017solving}. Two chemical reagents $A$ and $B$ are carried by the fluid flow, i.e., \textit{advected}, around the the cylinder and react to product chemical $C$, which is then disbursed downstream. The dynamical system of the chemical reactions is:
\begin{align*}
\frac{\partial u_A}{\partial t} + \mathbf{v} \cdot \nabla u_A - \nabla\cdot (\epsilon\nabla u_A) &= f_A - Ku_au_b\\
\frac{\partial u_B}{\partial t} + \mathbf{v} \cdot \nabla u_B - \nabla\cdot (\epsilon\nabla u_B) &= f_B - Ku_au_b\\
\frac{\partial u_C}{\partial t} + \mathbf{v} \cdot \nabla u_C - \nabla\cdot(\epsilon\nabla u_C) &= f_C + Ku_au_b - Ku_C\\
\end{align*}
The functions $u_A$, $u_B$, and $u_C$ represent the concentration of chemicals A, B, and C respectively at a given time and $xy$-coordinate. $f_A$ and $f_B$ represent the source terms for chemicals $A$ and $B$, which enter the domain at opposite sides (see Figures \ref{fig:diffusionreactiona} and  \ref{fig:diffusionreactionb}). The parameter $\epsilon=0.01$ corresponds to the domain diffusivity, while $K=10$ is the reaction rate.

\begin{figure}[h!]
    \centering
    \includegraphics[width=\linewidth]{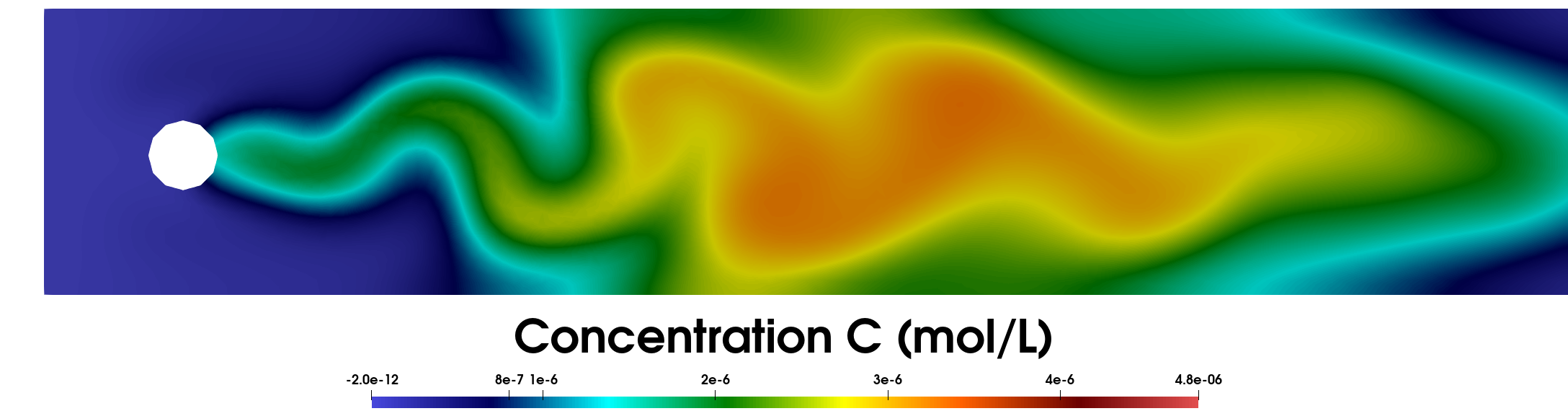}
    \caption{Concentration of \textit{C} in the \textit{Reaction-Diffusion Cylinder} system at $t=199$.}
    \label{fig:diffusionreactionc}
\end{figure}

The \textit{Reaction-Diffusion Cylinder} is simulated for 5.0 seconds (matching the \textit{NS Cylinder} velocity solution), saving the data at 100 $Hz$ for a total of 500 timesteps. The unstructured solution can be regenerated using the script \texttt{fenics\_diffusion\_reaction\_cylinder.py}, from which the structured version can be generated using the \texttt{diffusion\_reaction\_cylinder\_structured.py} script. We subsample the data both in time and space the same way as we did in the \textit{NS Cylinder} system.

\section{Methods and Hyperparameter Optimization}
This section provides a brief overview of the algorithms along with their corresponding hyperparameter sets. Tables \ref{tab:hyper-lm} and \ref{tab:hyper-other} list the hyperparameters used for each method.

\textbf{SINDy} \cite{brunton2016discovering} constructs a library of candidate functions (e.g., polynomials, trigonometric functions), evaluated on the state variables over time, and applies LASSO ($\ell_1$)-regularized regression to identify a sparse set of terms that best represent an ODE system. \textbf{PDEFind} \cite{rudy2017data} extends SINDy to PDEs by evaluating the candidate library on a spatiotemporal grid.

\textbf{WSINDy} \cite{messenger2021weak} is another linear method for PDE discovery that avoids explicit pointwise derivative estimation by leveraging the weak form of the governing equations. \textbf{ESINDy} and \textbf{EWSINDy} \cite{fasel2022ensemble} extend SINDy and WSINDy, respectively, using bootstrap aggregation for improved robustness on ODE and PDE systems. All SINDy-based methods are implemented using the  PySINDy library \url{https://pysindy.readthedocs.io/en/latest/}.

\textbf{DeepMoD} \cite{both2021deepmod} is a linear approach that integrates a neural network to simultaneously learn the underlying function, its derivatives, and the coefficients of a predefined candidate library. \textbf{Bayesian} \cite{more2023bayesian} builds on PDEFind by incorporating variational Bayes and a spike-and-slab prior for sparse model selection. Its implementation is available at \url{https://github.com/TapasTripura/Bayesian-Discovery-of-PDEs}.

\textbf{PySR} \cite{cranmer2023interpretable} is a genetic programming method with an optimized Julia backend, enabling high-performance model discovery through a parallel population-based evolve-simplify-optimize loop. See the official repository at \url{https://github.com/MilesCranmer/PySR}. \textbf{Operon} \cite{burlacu2020operon} is another efficient genetic programming method, implemented in C++ with vectorized operations and a compact linear tree representation. Its codebase is available at \url{https://github.com/heal-research/operon}.

\textbf{EQL} \cite{martius2016extrapolation} is a deep learning-based method that embeds symbolic operations (e.g., addition, cosine) into the activation functions of a shallow neural network. After training, the network weights correspond to the coefficients of the discovered equation. We use the implementation available at \url{https://github.com/cavalab/eql}. \textbf{uDSR} \cite{landajuela2022unified} provides a unified framework that integrates multiple approaches, from linear models to genetic programming, for model discovery. Its code is available at \url{https://github.com/dso-org/deep-symbolic-optimization}.

\textbf{End2end} \cite{kamienny2022end} is a general-purpose model discovery approach based on a sequence-to-sequence Transformer architecture, pretrained on large datasets of synthetic equations. The implementation can be found at \url{https://github.com/facebookresearch/symbolicregression}. \textbf{ODEFormer} \cite{d2024odeformer} extends End2end to ODE systems. Unlike most generic model discovery methods, ODEFormer does not rely on derivative estimates and instead takes state trajectories and timestamps as input. Its implementation is available at \url{https://github.com/sdascoli/odeformer}.

\begin{table*}[ht]
\centering
\begin{tabular}{l|l|l}
\toprule
\textbf{Model} & \textbf{Hyperparameter} & \textbf{Values} \\
\hline
\textbf{PDEFIND} & threshold & \texttt{np.logspace(-7, 0, 16)} \\
                 & basis functions & [polynomial], [polynomial, sin, cos] \\
                 & derivative order & 1, 2, 3, 4 \\
                 & polynomial order & 1, 2, 3, 4 \\
                 & alpha & $10^{-5}$, $10^{-4}$ \\
                 & optimizer & SLTSQ \\
                 & max iterations & 200 \\
\hline
\textbf{WSINDy} & threshold & \texttt{np.logspace(-7, 0, 16)} \\
                 & basis functions & [polynomial], [polynomial, sin, cos] \\
                 & derivative order & 1, 2, 3, 4 \\
                 & polynomial order & 1, 2, 3, 4 \\
                 & integration points & 200, 2000 \\
                 & optimizer & SR3 \\
                 & max iterations & 200 \\
\hline
\textbf{E(W)SINDy}  & \#models & [10, 20, 50] \\
(In addition to SINDy & subset ratio & [0.5, 0.7, 0.9]  \\
and WSINDy hyperparamters) & inclusion prob. threshold & [0.2, 0.3, 0.4, 0.5] \\
\hline
\textbf{Bayesian} & threshold & \texttt{np.logspace(-7, 0, 16)} \\
                 & basis functions & [polynomial] \\
                 & derivative order & 1, 2, 3, 4 \\
                 & polynomial order & 1, 2, 3, 4 \\
                 & alpha & $10^{-5}$, $10^{-4}$ \\
                 & tolerance & $10^{-4}$, $10^{-3}$, $10^{-2}$ \\
                 & pip & 0.5, 0.7, 0.9 \\
\hline
\textbf{DeepMoD} & derivative order & 1, 2, 3, 4 \\
                 & polynomial order & 1, 2, 3, 4 \\
                 & hidden layers    & [[50, 50, 50, 50]] \\
                 & learning rate    & 0.001 \\
                 & threshold        & [0.1, 0.3, 0.5] \\
\hline
\textbf{SINDy} & threshold & \texttt{np.linspace(0.001, 1, 10)} \\
                 & basis functions & \makecell[l]{{[polynomial], [polynomial, sin, cos]}\\{[polynomial, sin, cos, exp]}} \\
                 & polynomial order & 1, 2, 3, 4 \\
                 & alpha & 0.025, 0.05, 0.075 \\
                 & optimizer & SLTSQ \\
                 & max iterations & 20, 50, 100 \\               
\bottomrule
\end{tabular}
\caption{Hyperparameter settings used for LM discovery methods.}
\label{tab:hyper-lm}
\end{table*}

\begin{table*}[ht]
\centering
\begin{tabular}{l|l|l}
\toprule
\textbf{Model} & \textbf{Hyperparameter} & \textbf{Values} \\
\hline
\textbf{EQL} & iterations & 10000 \\
            & regularization & 0.0001, 0.001, 0.01, 0.05 \\
            & \#layers   & [1, 2] \\
            & activation functions (layer 1) & [id;mul;cos;sin;exp;square;sqrt;id;mul;cos;sin;exp;square;sqrt;log]\\
            & activation functions (layer 2)& [id;mul;cos;div;sqrt;cos;sin;div;mul;mul;cos;id;log] \\
\hline
\textbf{ODEFormer}  & beam temperature  & [0.05, 0.1, 0.2, 0.3, 0.5] \\
                    & beam size         & 50 \\
\hline
\textbf{End2end}    & max input points  & 200 \\
                    & \#trees to refine& 100 \\
                    & rescale           & True \\
\hline
\textbf{PySR}       & \#iterations, \#cycles per iteration  & 100, 1000 \\
                    & \#populations, population size        & 20, 100 \\
                    & max size, max depth                   & 40, 20 \\
                    & binary operators                      & [[$+$, $-$, $\times$, $/$]] \\
                    & unary operators                       & [[sin, exp, log, sqrt]] \\
\hline
\textbf{Operon}     & symbols           & [[add, sub, mul, aq, sin, constant, variable]] \\
                    & brood size        & 10 \\
                    & max depth, max length & 10, 50 \\
                    & pool size, population size  & 1000, 1000 \\
\hline
\textbf{uDSR}       & function set           & [[add, sub, mul, div, sin, cos, exp, poly, const]] \\
                    & poly optimizer params degree & 3 \\
                    & poly optimizer params coef tol & $10^{-8}$ \\
                    & gp-meld population size & 100 \\
                    & gp-meld generations & 20 \\
                    & gp-meld p-crossover &  0.5 \\
                    & gp-meld p-mutate & 0.5 \\
                    & training n samples & 100000  \\
                    & training batch size & 1000 \\
                    & training epsilon & 0.05 \\
                    & policy optimizer learning rate & 0.0005 \\
                    & policy optimizer entropy weight & 0.05 \\
                    & policy optimizer entropy gamma & 0.7 \\
\bottomrule
\end{tabular}
\caption{Hyperparameter settings used for GP, DL, and LSPT model discovery methods.}
\label{tab:hyper-other}
\end{table*}


\section{Additional Experimental Results}
This section provides more detailed results on the runtime and performance of the the evaluated methods on ODE and PDE systems.

\subsection{ODE Systems}
\begin{figure}[!ht]
\centering
\begin{subfigure}{\linewidth}
    \centering
    \includegraphics[width=\linewidth]{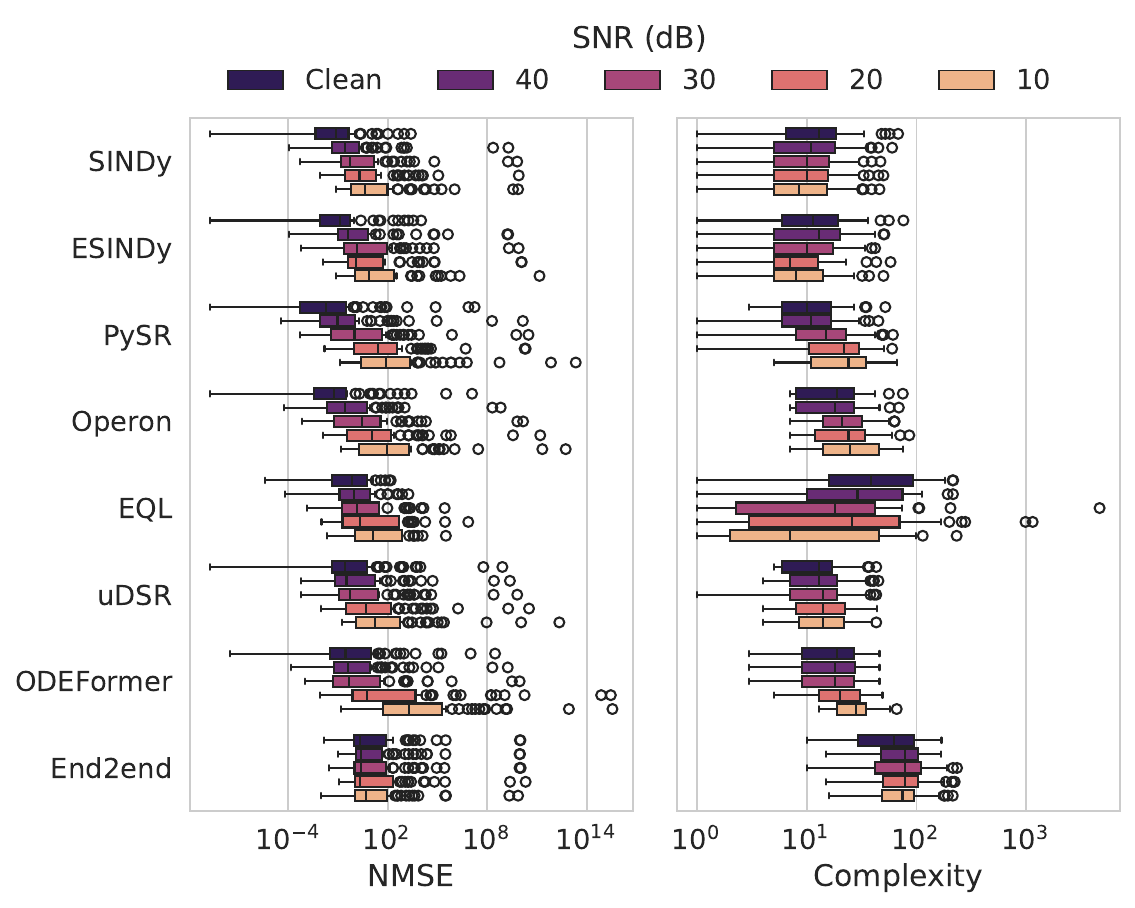}
    \caption{Performance of the model discovery methods on ODE datasets.}
    \label{fig:perf-ode-supp}
\end{subfigure}
\begin{subfigure}{\linewidth}
    \centering
    \includegraphics[width=\linewidth]{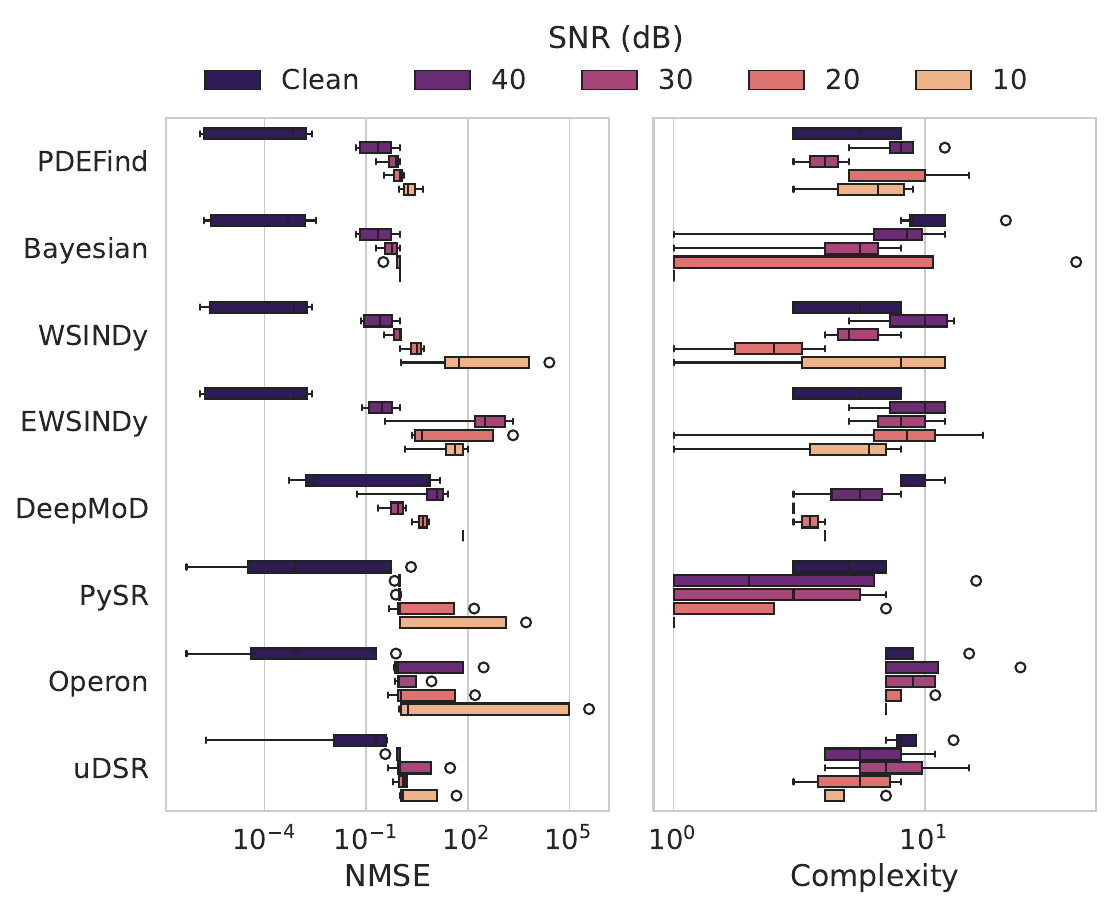}
    \caption{Performance of the model discovery methods on a selected subset of PDE datasets.}
    \label{fig:perf-pde-supp}
\end{subfigure}

\caption{Box plots of model discovery performance on ODE and PDE datasets. Although PDE systems are generally more complex than ODEs, the observed errors are often higher for ODEs. This is primarily because, in some ODE datasets, the test trajectories lie in regions of stable dynamics where the true derivatives are near zero. As a result, even small absolute errors can lead to large NMSE values, despite the discovered equations being reasonably accurate.}
\end{figure}

\begin{figure*}[h]
    \centering
    \begin{subfigure}{0.49\textwidth}
        \includegraphics[width=\linewidth]{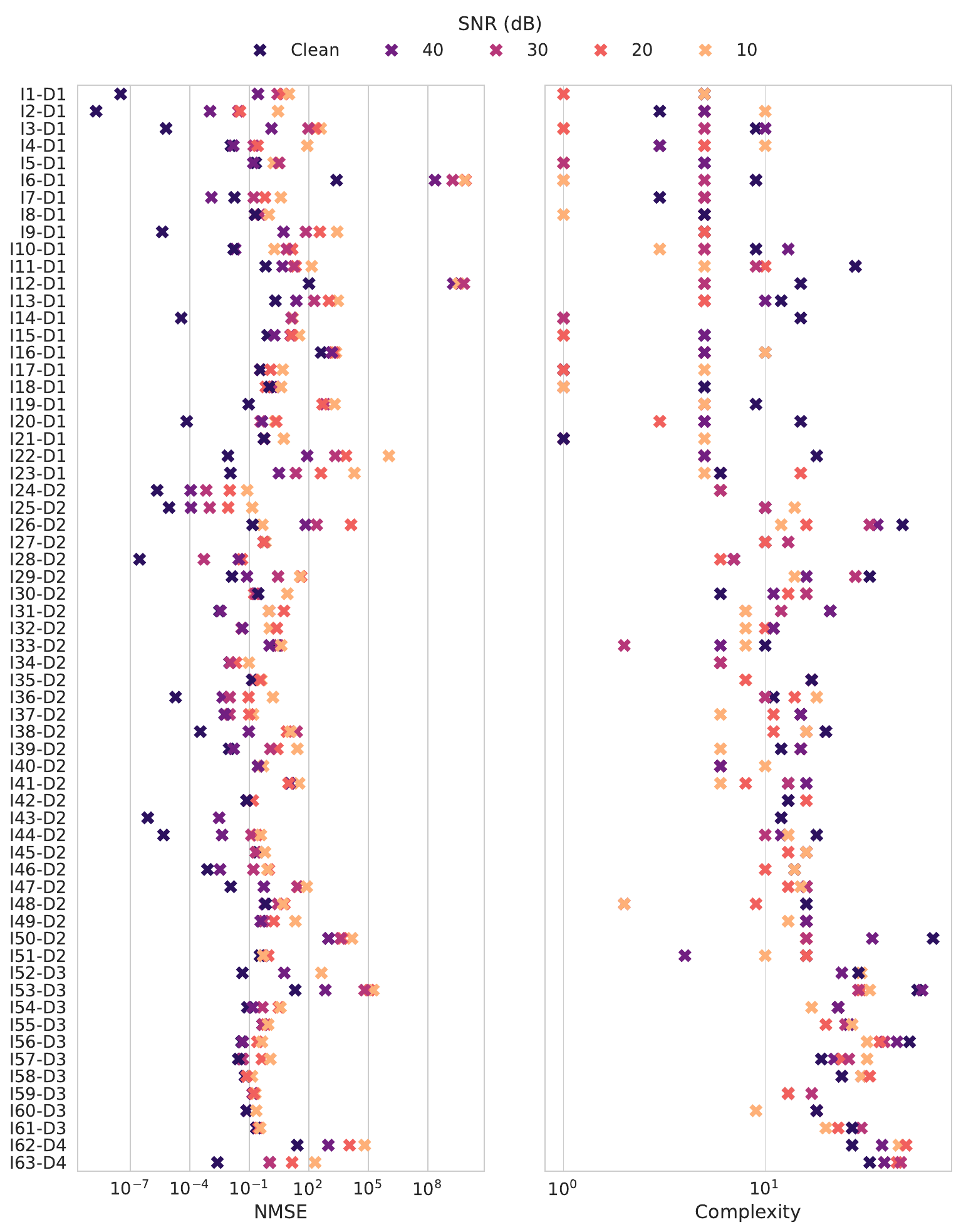}
        \caption{SINDy}
    \end{subfigure}
    \hfill
    \begin{subfigure}{0.49\textwidth}
        \includegraphics[width=\linewidth]{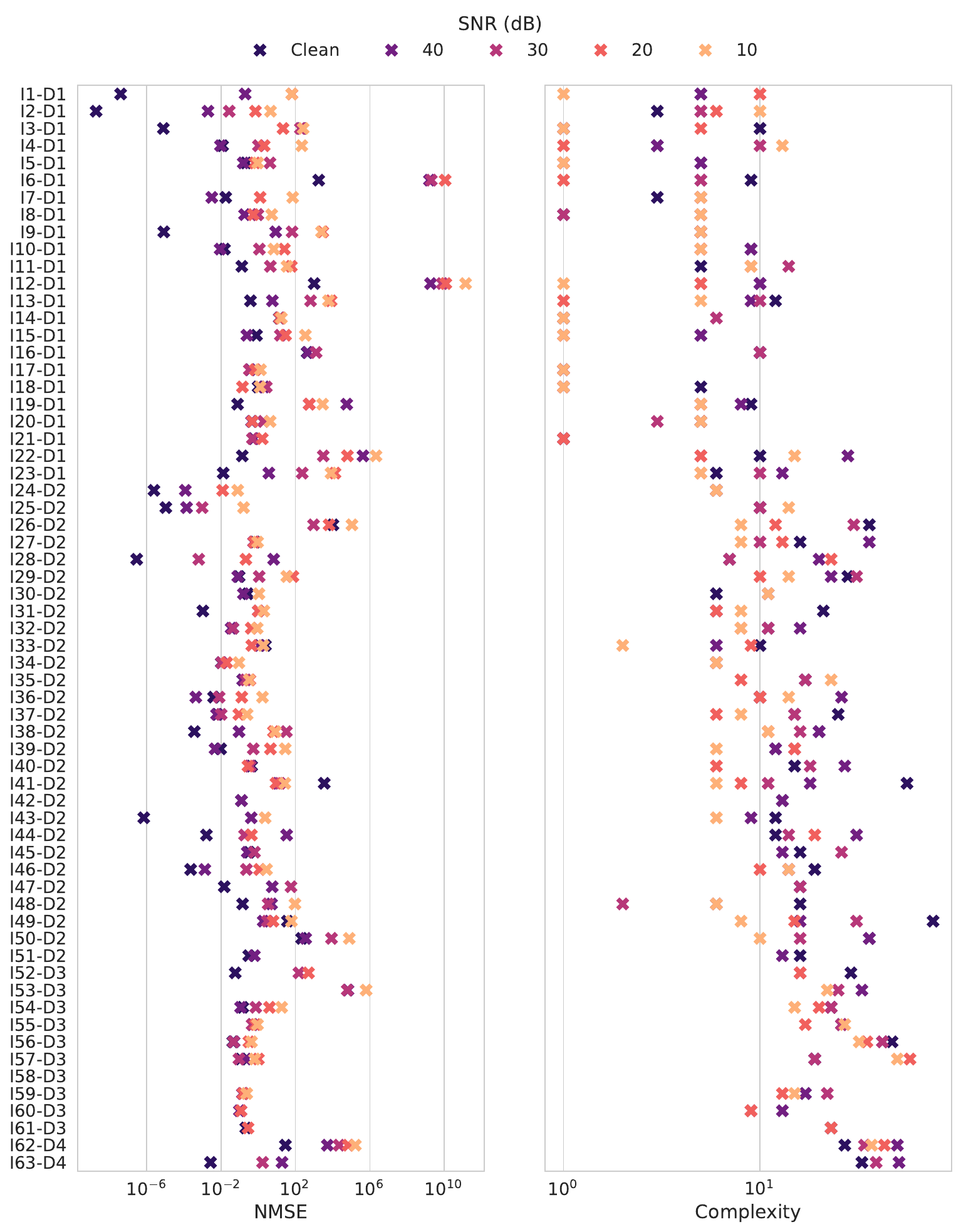}
        \caption{ESINDy}
    \end{subfigure}
    \begin{subfigure}{0.49\textwidth}
        \includegraphics[width=\linewidth]{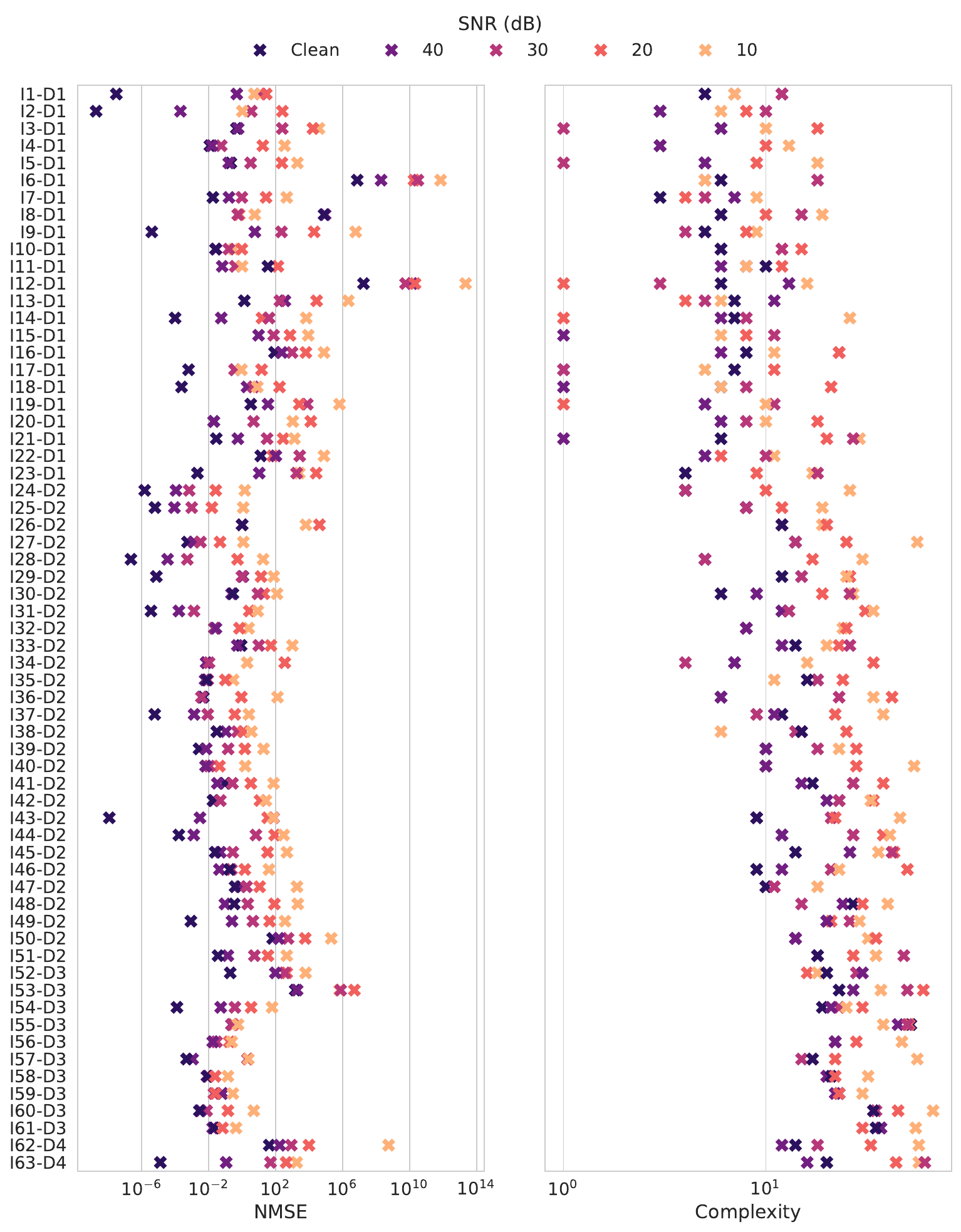}
        \caption{PySR}
    \end{subfigure}
    \hfill
    \begin{subfigure}{0.49\textwidth}
        \includegraphics[width=\linewidth]{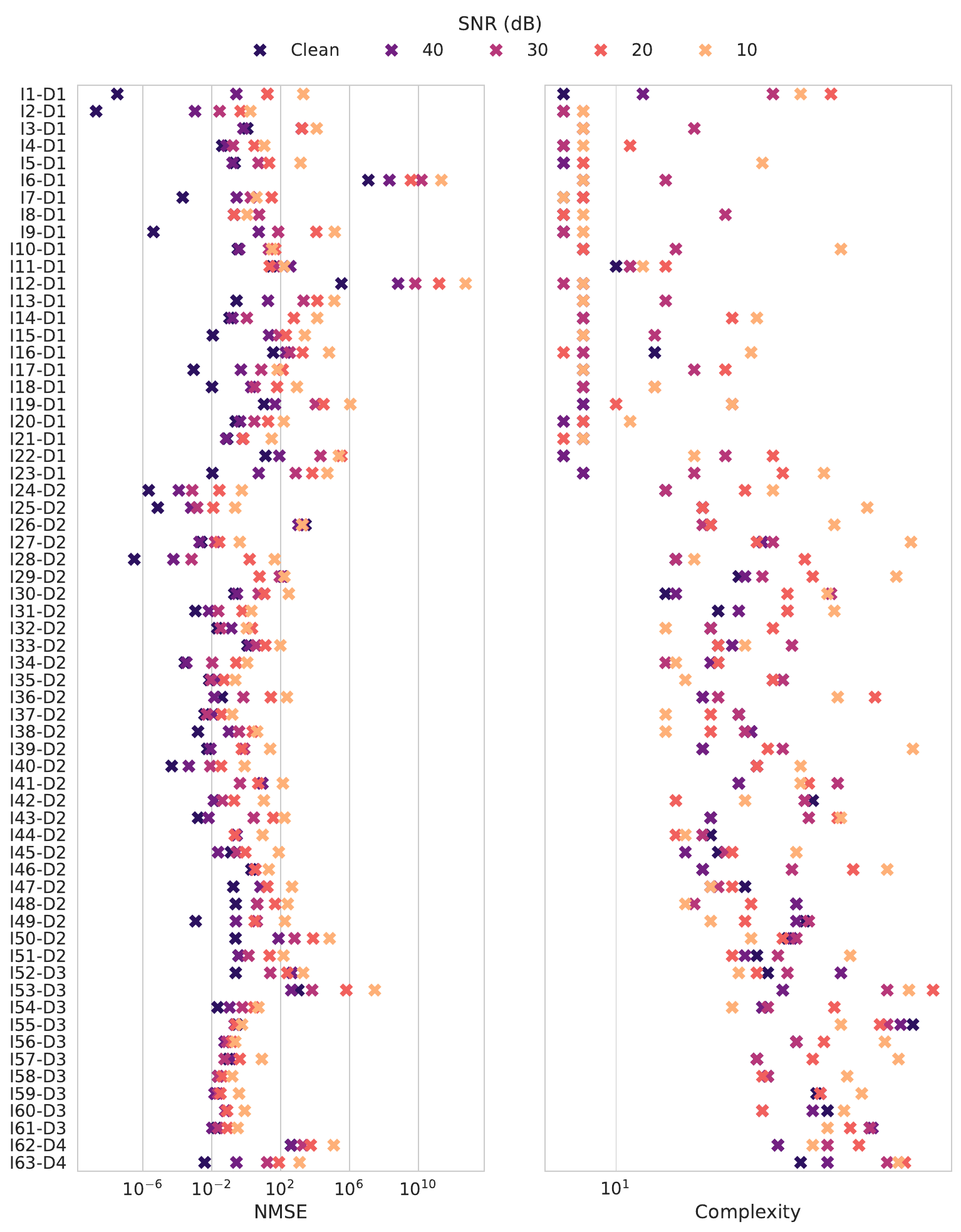}
        \caption{Operon}
    \end{subfigure}
    \caption{Performance of SINDy, ESINDy, PySR, and Operon on ODE datasets: complexity of the discovered equations and the NMSE of time derivatives on the test set.}
    \label{fig:perf-ode-part1}
\end{figure*}

\begin{figure*}[h]
    \centering
    \begin{subfigure}{0.49\textwidth}
        \includegraphics[width=\linewidth]{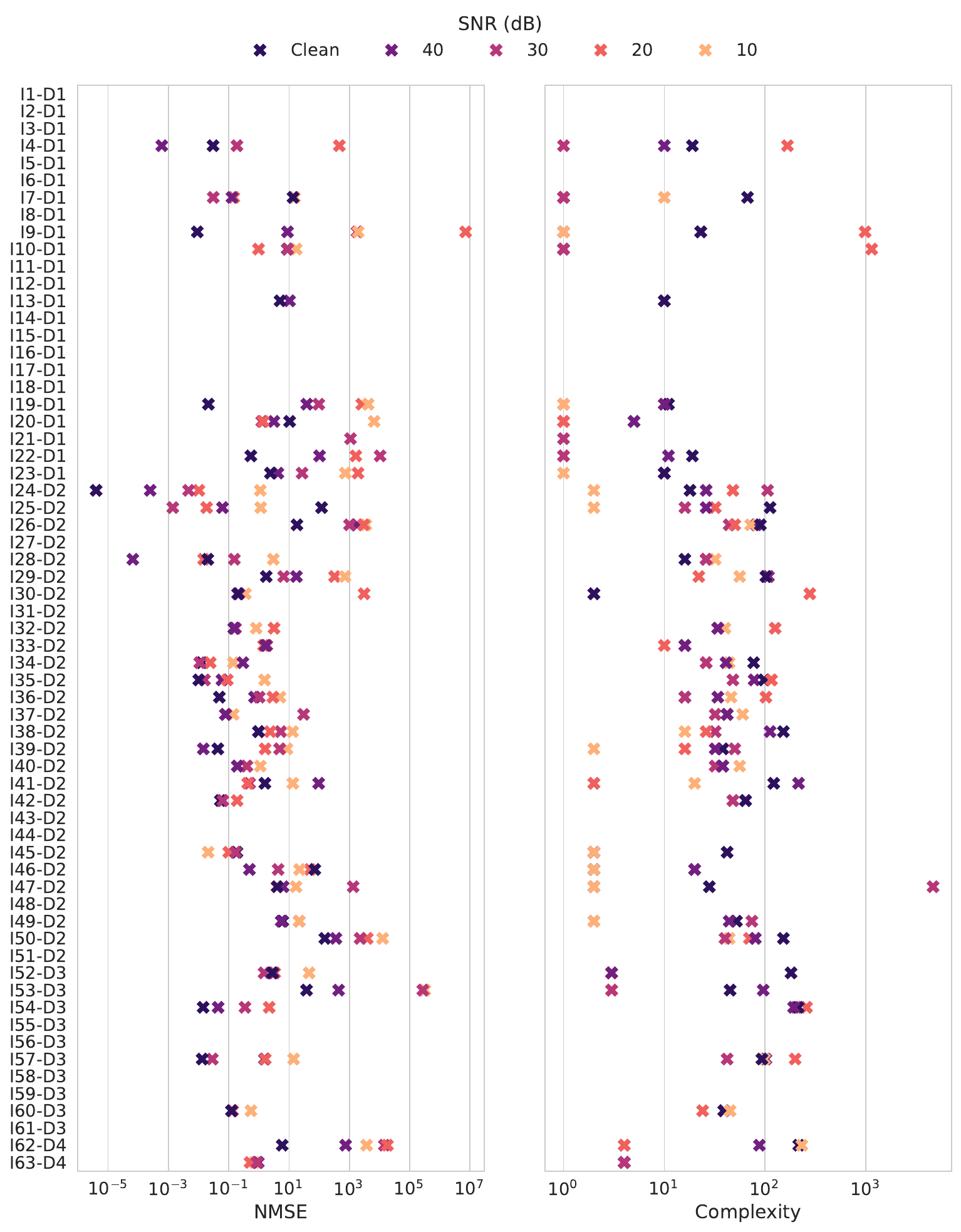}
        \caption{EQL}
    \end{subfigure}
    \hfill
    \begin{subfigure}{0.49\textwidth}
        \includegraphics[width=\linewidth]{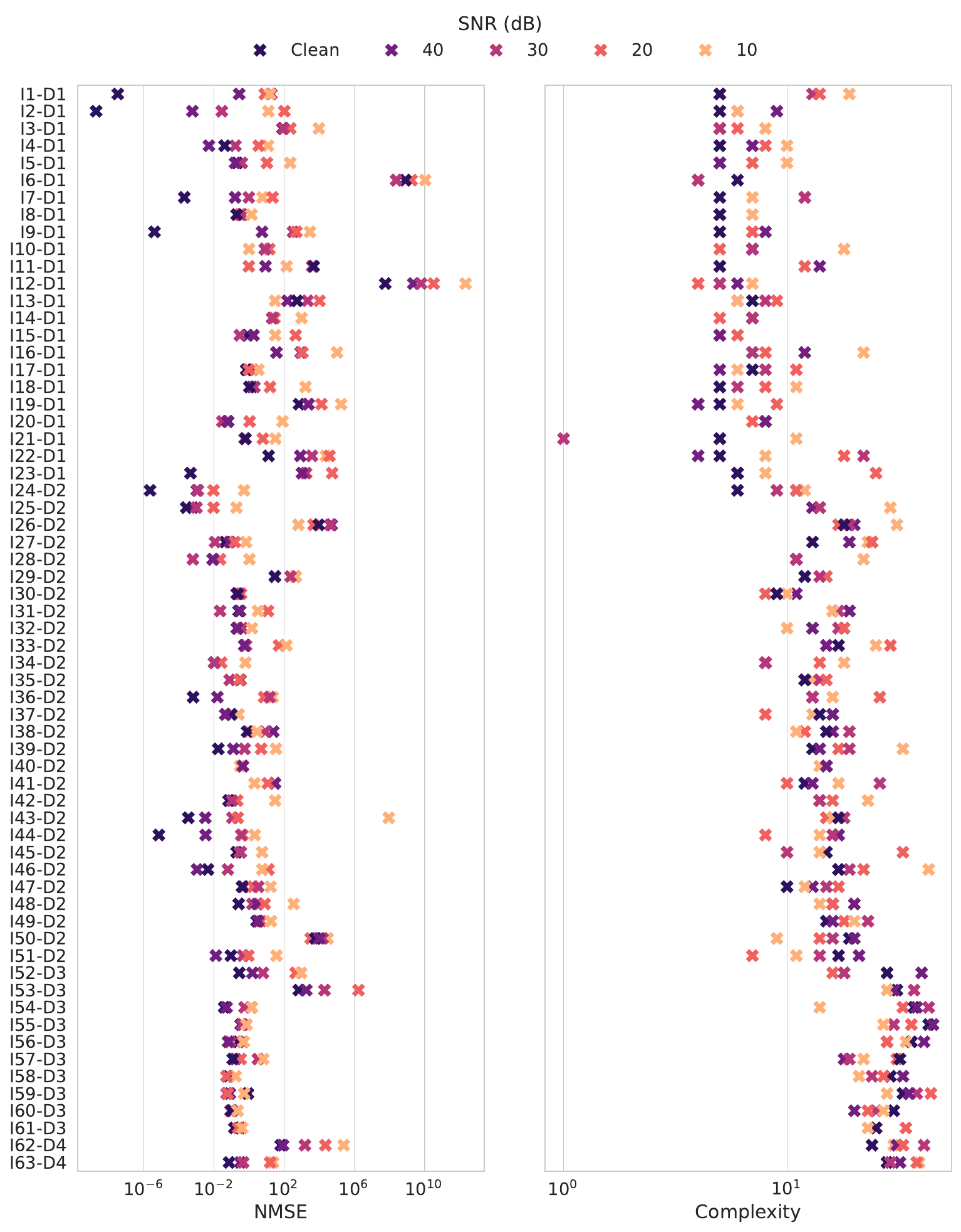}
        \caption{uDSR}
    \end{subfigure}
    \begin{subfigure}{0.49\textwidth}
        \includegraphics[width=\linewidth]{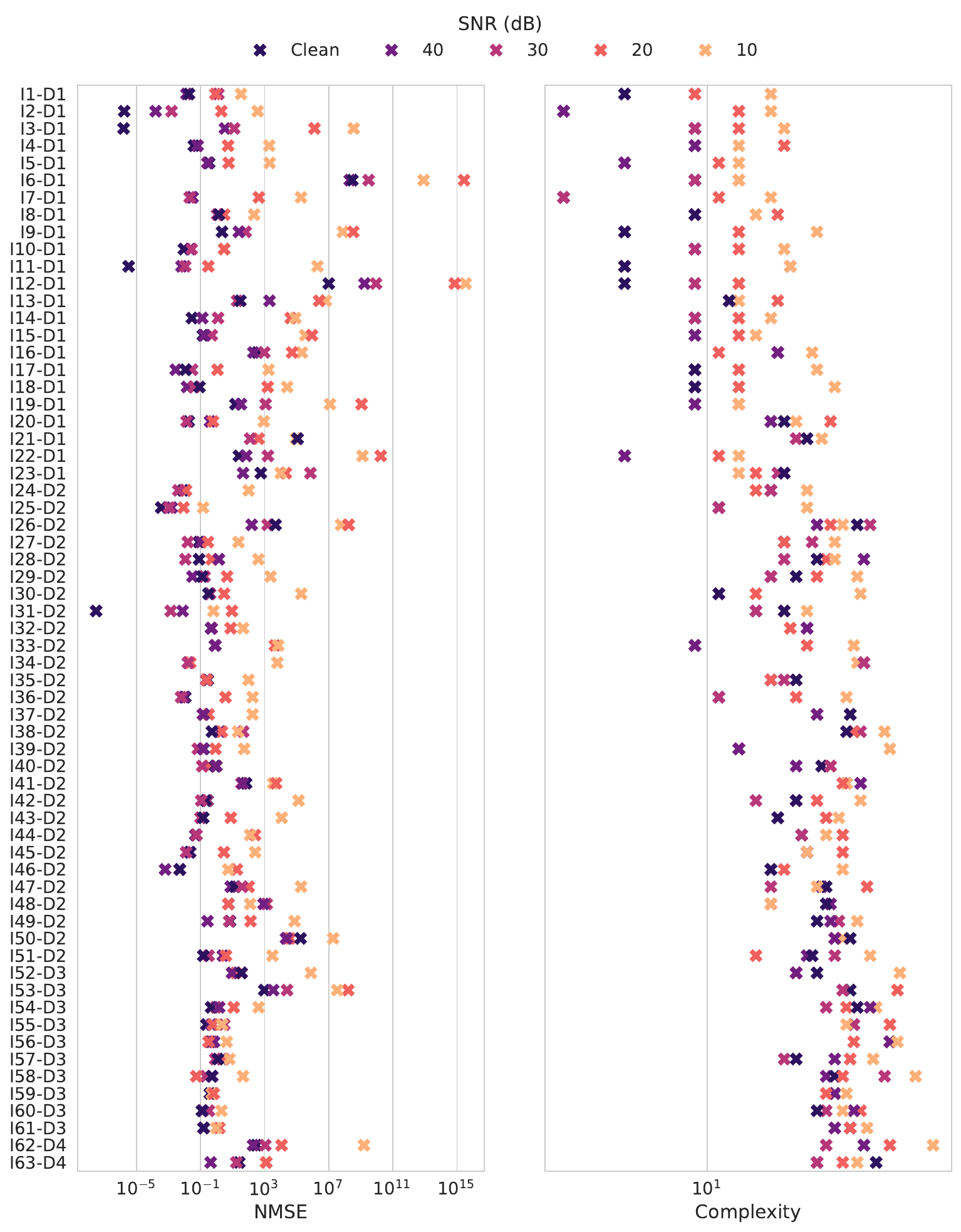}
        \caption{ODEFormer}
    \end{subfigure}
    \hfill
    \begin{subfigure}{0.49\textwidth}
        \includegraphics[width=\linewidth]{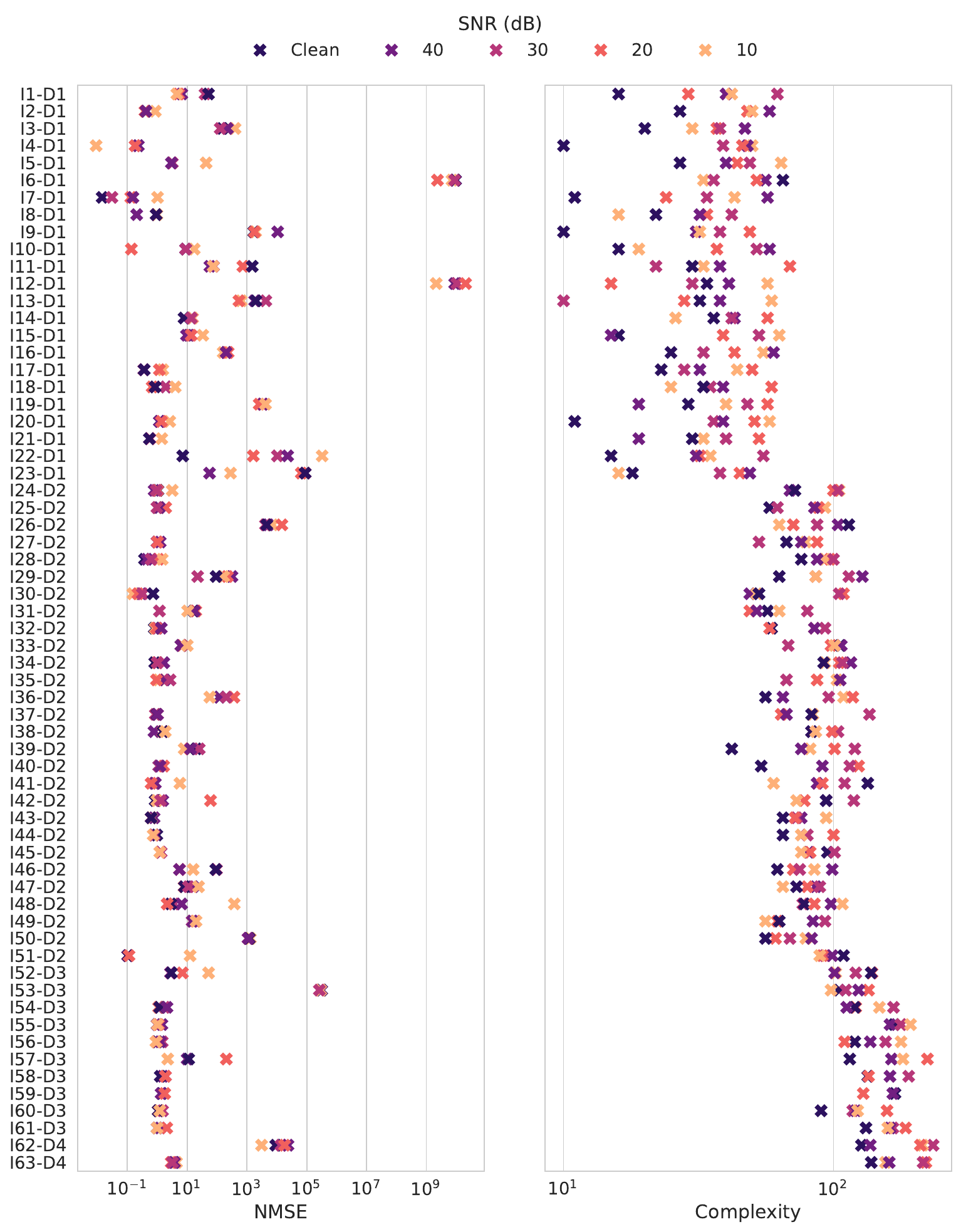}
        \caption{End2end}
    \end{subfigure}
    \caption{Performance of EQL, uDSR, ODEFormer, and End2end on ODE datasets: complexity of the discovered equations and the NMSE of time derivatives on the test set.}
    \label{fig:perf-ode-part2}
\end{figure*}

Figure \ref{fig:perf-ode-supp} presents aggregated performance of model discovery methods on all ODE systems, as measured by NMSE of predicted derivatives on the test set. While LM tend to discover less complex equations as the noise level increases, GP methods and ODEFormer exhibit the opposite trend. Figures \ref{fig:perf-ode-part1} and \ref{fig:perf-ode-part2} provide more detailed metrics on how different noise levels affect the quality of equations for each dataset. The dataset names are described in Table \ref{tab:ode-dataset-description}. For more details about the equations and initial values, refer to \cite{d2024odeformer} or \texttt{scripts/strogatz\_ode.py} in the MDBench project repository.

As we discuss in the Section \ref{sec:limitations}, no established equation similarity metric exists to enable automatic assessment of equation fidelity. Therefore, in order to have a closer look at the fidelity of generated equations to the ground truth dynamics, we manually compare the discovered and true equations for ODE datasets in noise-free settings. Tables \ref{tab:ode-performance-1} and \ref{tab:ode-performance-2} show the fidelity of the discovered equations along with their NMSE and complexity on ODE systems. 

LM and GP methods have a higher successful discovery rate compared to other evaluated methods. It can be observed that EQL is the most unstable method with the highest rate of failure. Both EQL and End2end were not able to accurately discover any of the ODE systems in the clean settings. The inaccuracy of End2end method likely stems from the lack of dynamical system examples in the datasets that the underlying model was pretrained on. On the other hand, ODEFormer, while having a similar architecture to End2end, achieves a higher successful rediscovery rate. This suggests the necessity of including representative systems in the pretraining datasets for LSPT methods. A promising direction for improving the LSPT methods is to extend End2end and ODEFormer architectures to PDE systems, where a single transformer model is jointly trained on synthetic ODE and PDE systems.

\begin{figure}
    \centering
    \includegraphics[width=1\linewidth]{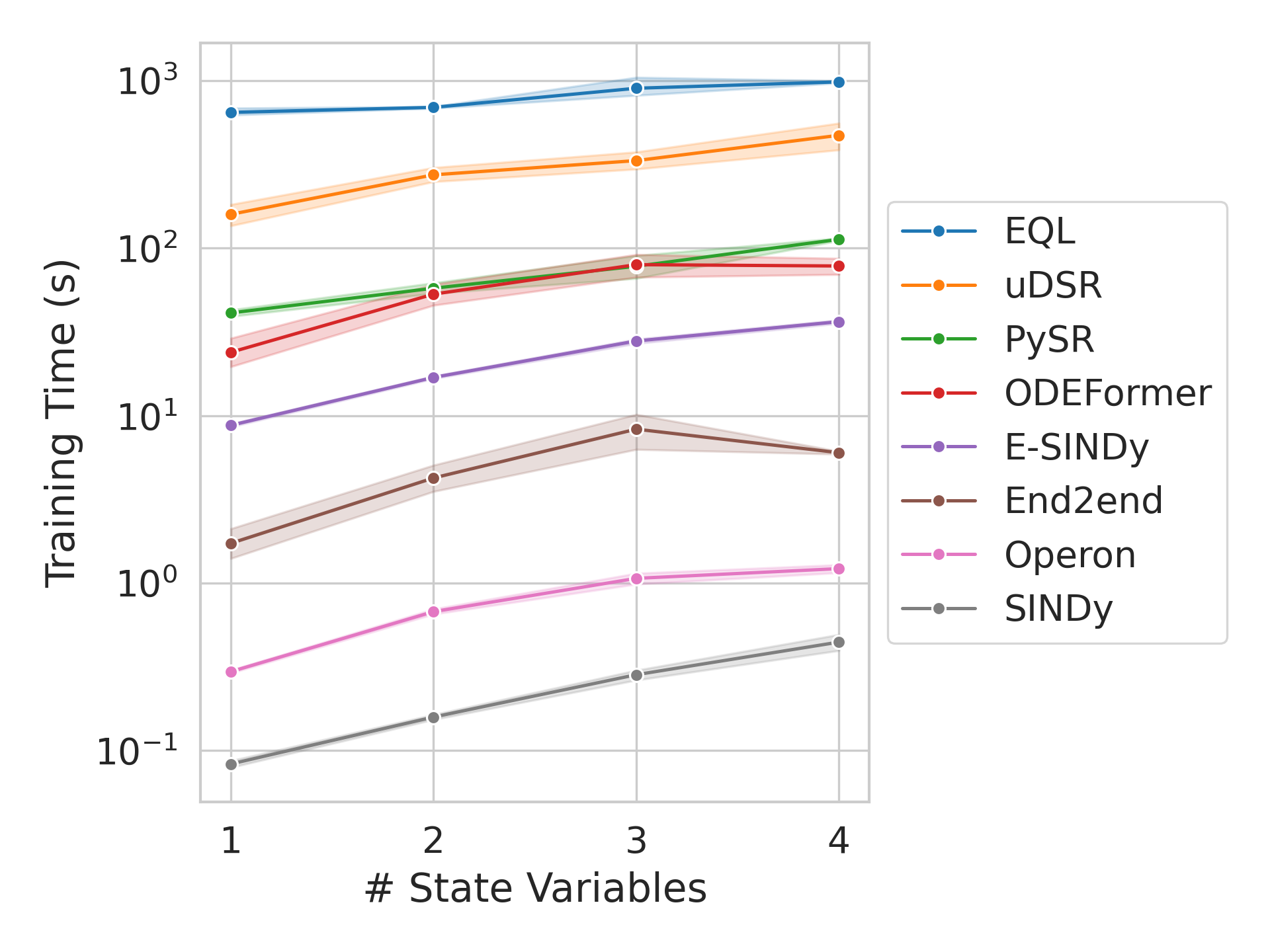}
    \caption{Training time for ODE datasets.}
    \label{fig:ode-train-time}
\end{figure}

Figure \ref{fig:ode-train-time} shows the average and 95\% confidence interval for training time of each method on ODE datasets across different noise levels. The DL based methods - EQL and uDSR - possess the longest training time, whereas SINDy has the fastest execution time, taking less than a second for each ODE dataset.

\subsection{PDE Systems}
\begin{figure*}[h]
    \centering
    \begin{subfigure}{0.49\textwidth}
        \includegraphics[width=\linewidth]{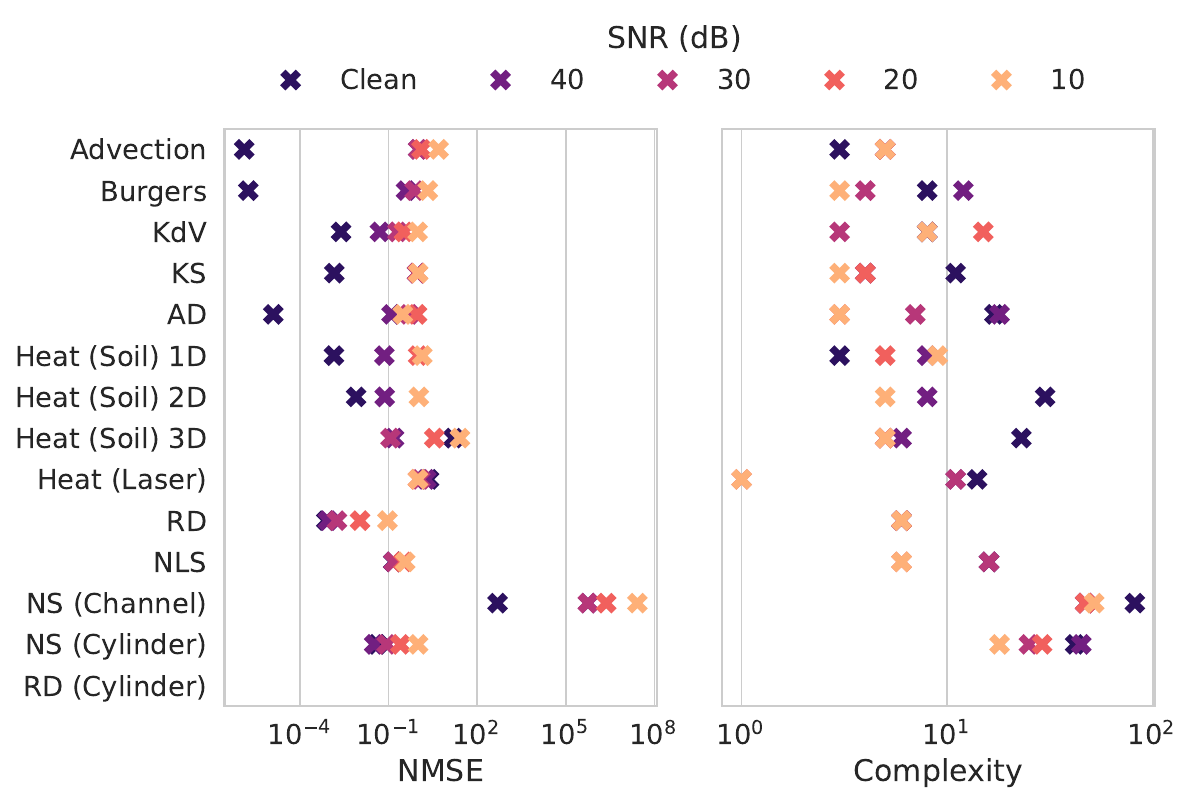}
        \caption{PDEFind}
    \end{subfigure}
    \hfill
    \begin{subfigure}{0.49\textwidth}
        \includegraphics[width=\linewidth]{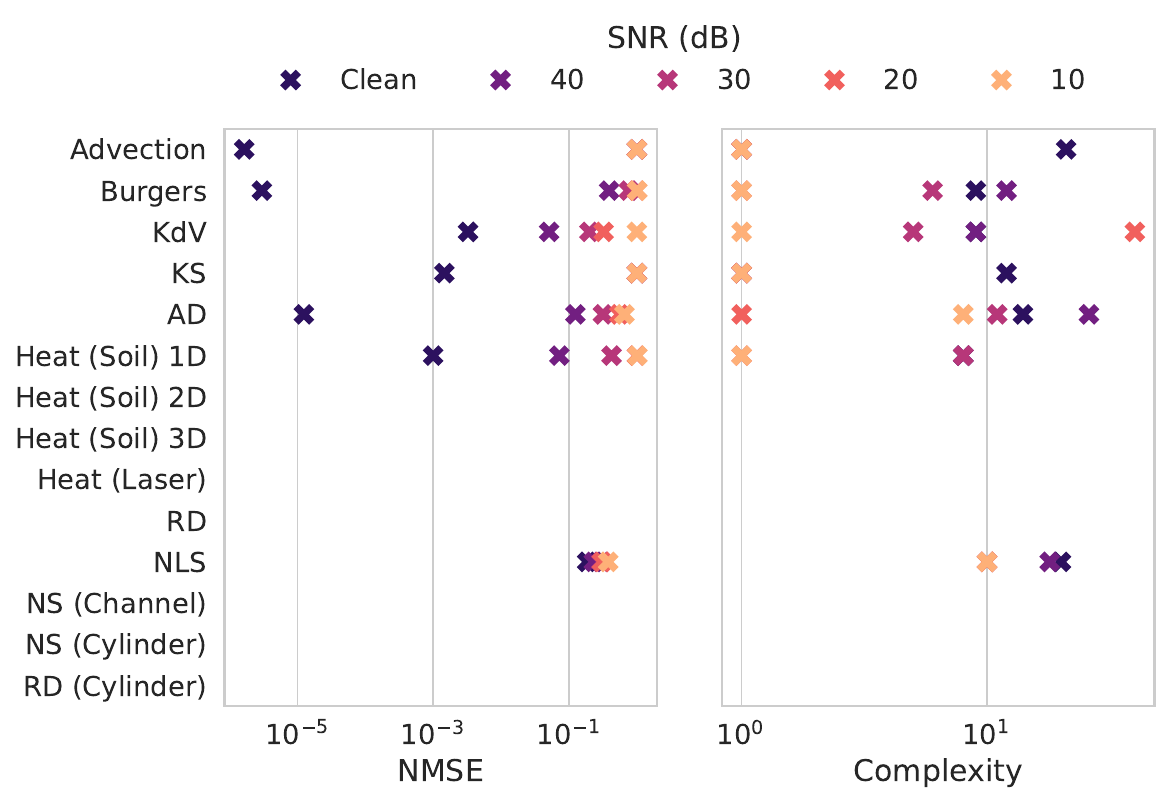}
        \caption{Bayesian}
    \end{subfigure}
    \begin{subfigure}{0.49\textwidth}
        \includegraphics[width=\linewidth]{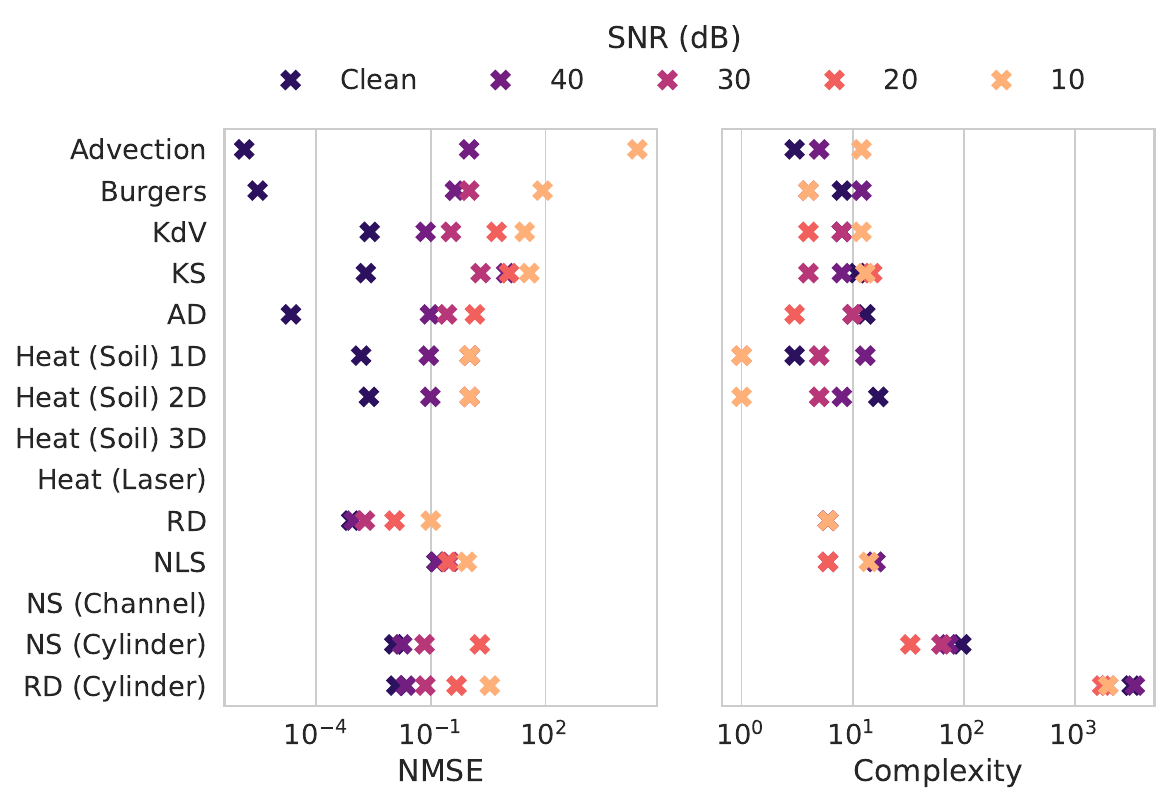}
        \caption{WSINDy}
    \end{subfigure}
    \hfill
    \begin{subfigure}{0.49\textwidth}
        \includegraphics[width=\linewidth]{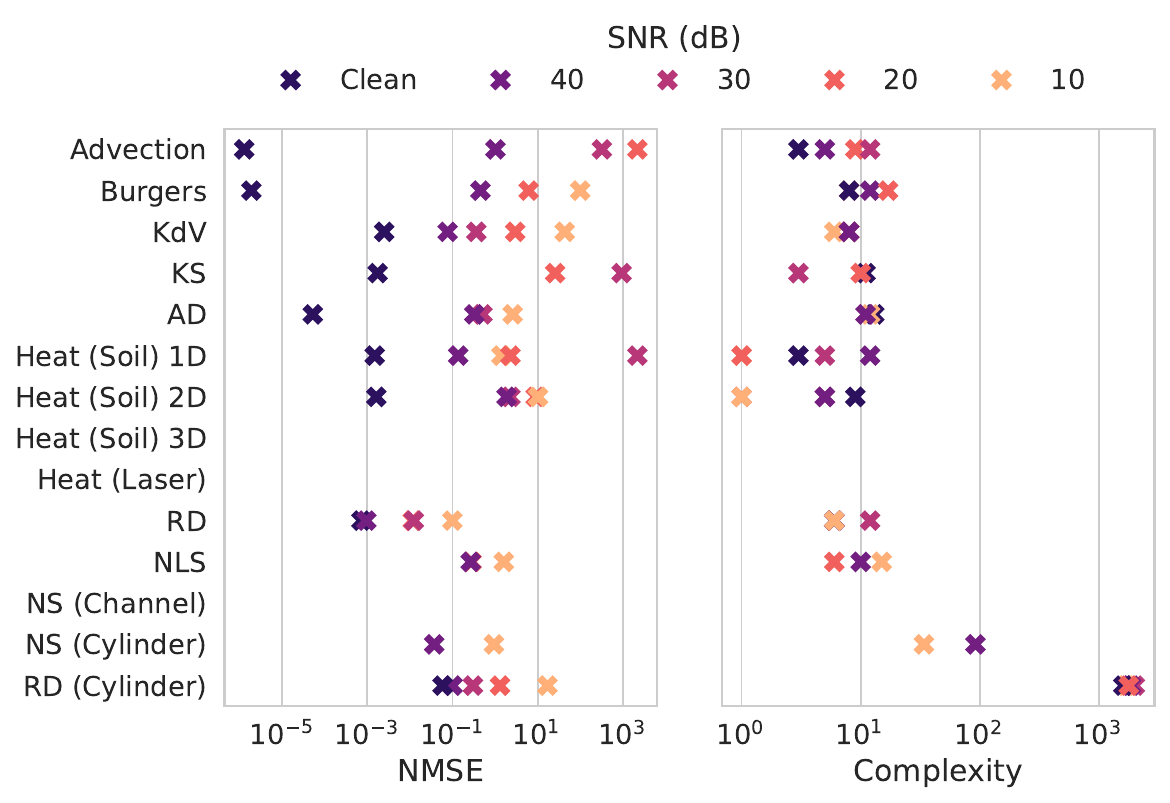}
        \caption{EWSINDy}
    \end{subfigure}
    \caption{Performance of PDEFind, Bayesian, WSINDy, and EWSINDy on PDE datasets: complexity of the discovered equations and the NMSE of time derivatives on the test set.}
    \label{fig:perf-pde-part1}
\end{figure*}

\begin{figure*}[h]
    \centering
    \begin{subfigure}{0.49\textwidth}
        \includegraphics[width=\linewidth]{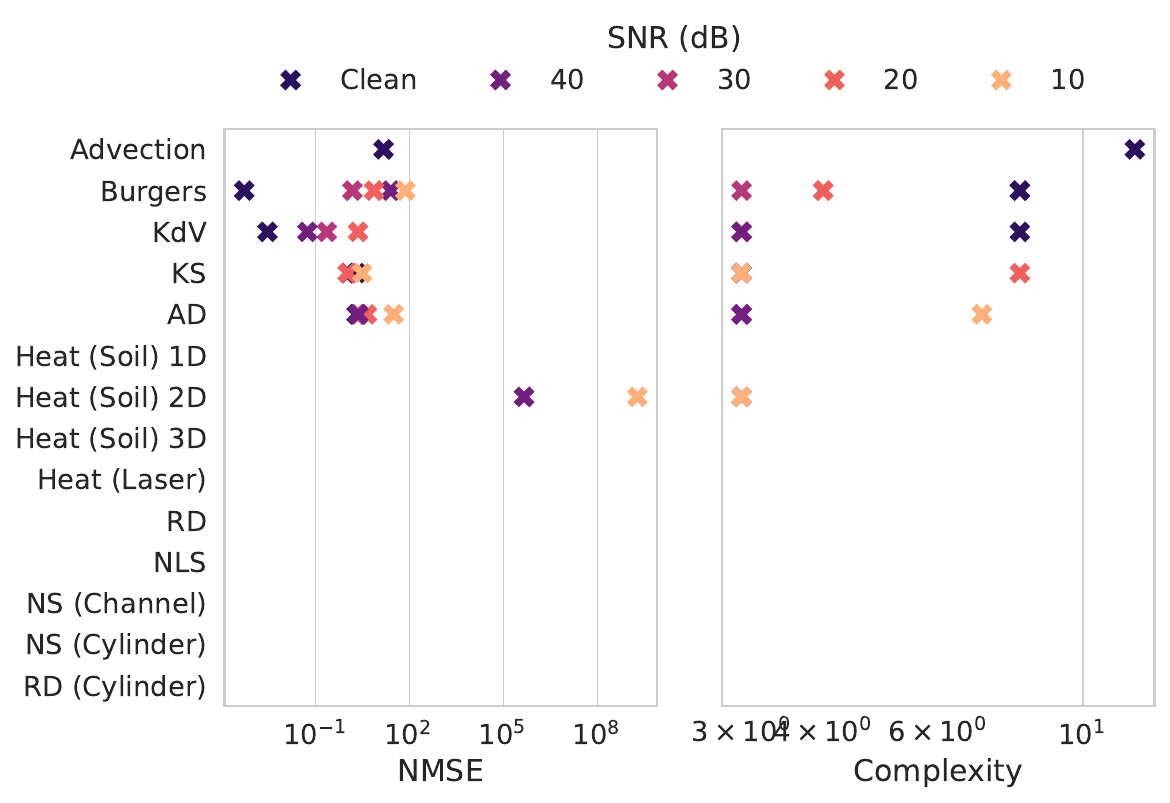}
        \caption{DeepMoD}
    \end{subfigure}
    \hfill
    \begin{subfigure}{0.49\textwidth}
        \includegraphics[width=\linewidth]{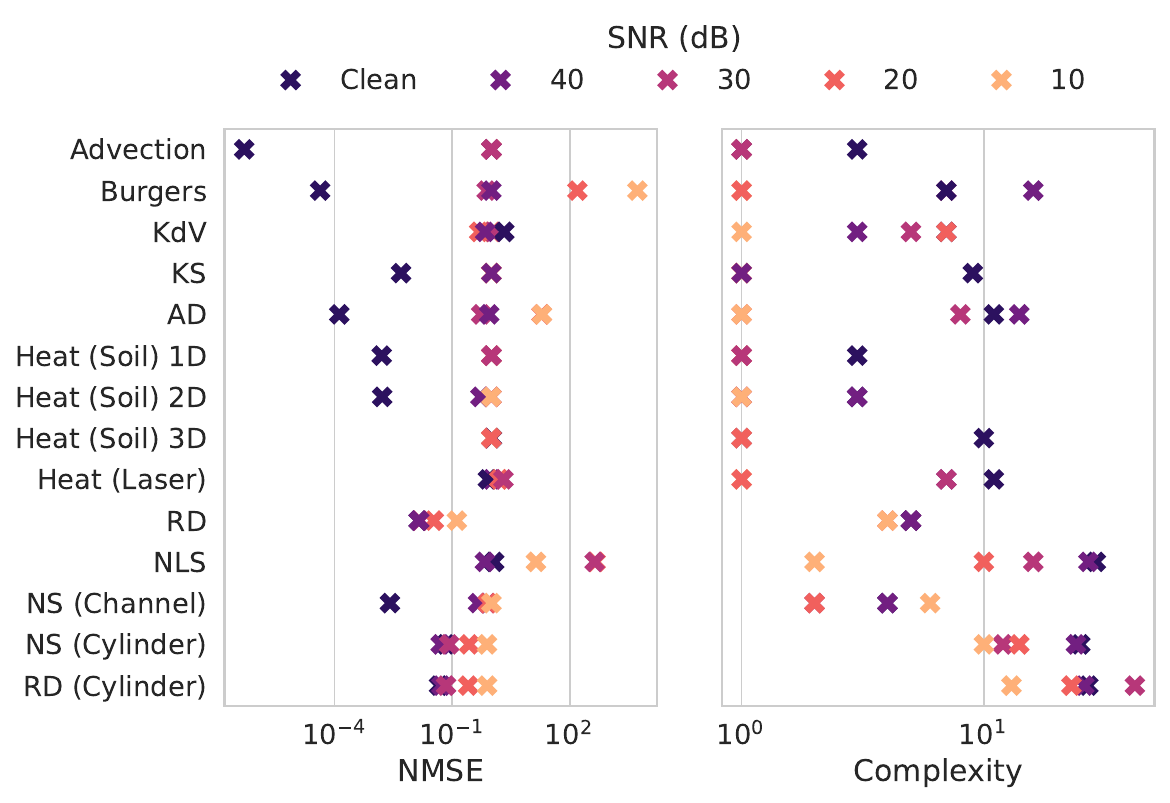}
        \caption{PySR}
    \end{subfigure}
    \begin{subfigure}{0.49\textwidth}
        \includegraphics[width=\linewidth]{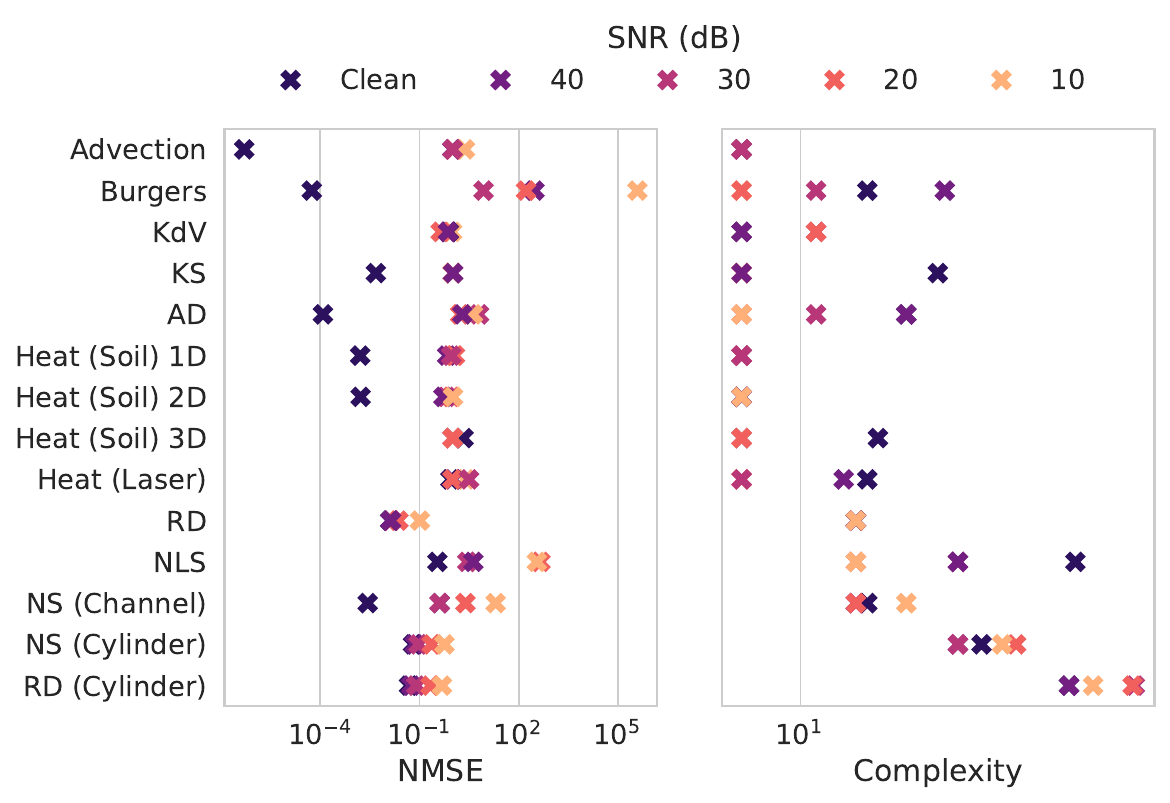}
        \caption{Operon}
    \end{subfigure}
    \hfill
    \begin{subfigure}{0.49\textwidth}
        \includegraphics[width=\linewidth]{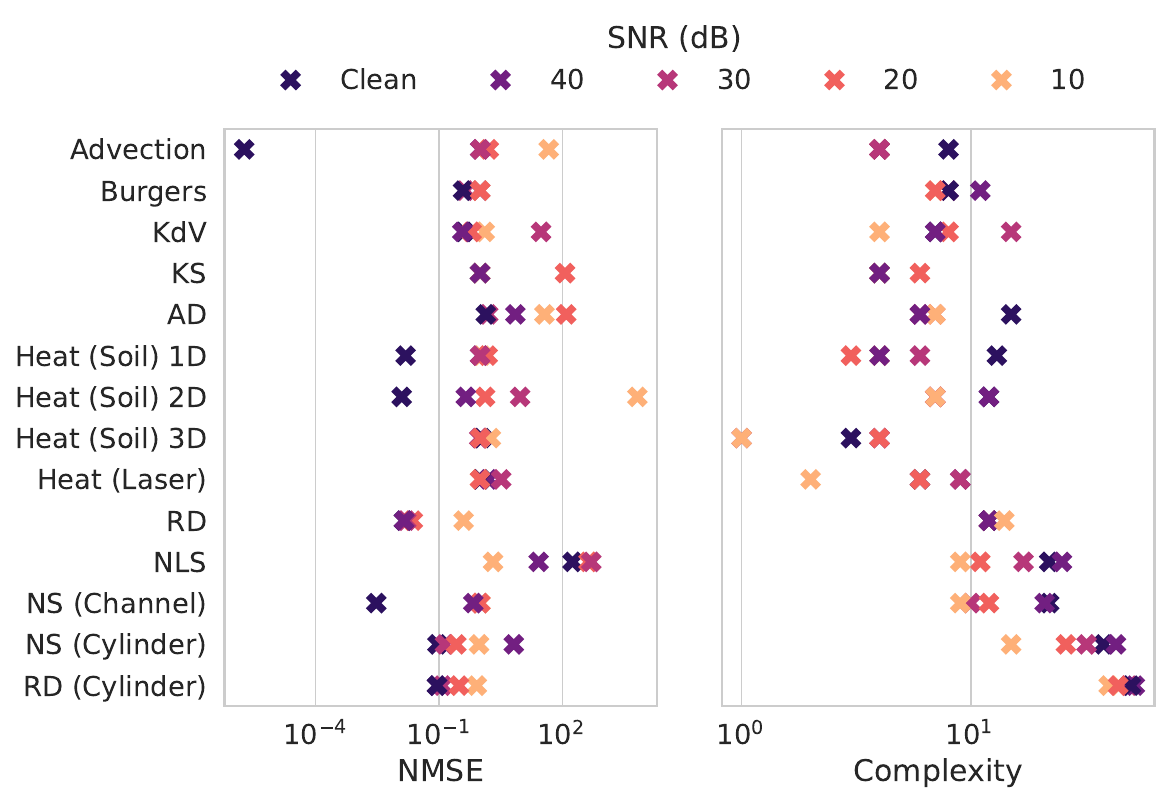}
        \caption{uDSR}
    \end{subfigure}
    \caption{Performance of DeepMoD, PySR, Operon, and uDSR on PDE datasets: complexity of the discovered equations and the NMSE of time derivatives on the test set.}
    \label{fig:perf-pde-part2}
\end{figure*}

Figures \ref{fig:perf-pde-part1} and \ref{fig:perf-pde-part2} presents the NMSE and complexity of discovered equations on each PDE dataset across varying noise levels. Generally, LM methods such as WSINDy achieved a lower prediction error compared to GP methods. The training runtimes on the aforementioned datasets are provided in Table \ref{tab:pde-times}. 

\begin{table*}[]
\small
    \centering
    \begin{tabular}{|cp{7cm}|cp{7cm}|}
    \toprule
    \textbf{Id} & \textbf{Equation Description} & \textbf{Id} & \textbf{Equation Description} \\
    \midrule
    I1-D1 & RC-circuit (charging capacitor) & I33-D2 & Glider (dimensionless) \\
I2-D1 & Population growth (naive) & I34-D2 & Frictionless bead on a rotating hoop (dimensionless) \\
I3-D1 & Population growth with carrying capacity & I35-D2 & Rotational dynamics of an object in a shear flow \\
I4-D1 & RC-circuit with non-linear resistor (charging capacitor) & I36-D2 & Pendulum with non-linear damping, no driving (dimensionless) \\
I5-D1 & Velocity of a falling object with air resistance & I37-D2 & Van der Pol oscillator (standard form) \\
I6-D1 & Autocatalysis with one fixed abundant chemical & I38-D2 & Van der Pol oscillator (simplified form from Strogatz) \\
I7-D1 & Gompertz law for tumor growth & I39-D2 & Glycolytic oscillator, e.g., ADP and F6P in yeast (dimensionless) \\
I8-D1 & Logistic equation with Allee effect & I40-D2 & Duffing equation (weakly non-linear oscillation) \\
I9-D1 & Language death model for two languages & I41-D2 & Cell cycle model by Tyson for interaction between protein cdc2 and cyclin (dimensionless) \\
I10-D1 & Refined language death model for two languages & I42-D2 & Reduced model for chlorine dioxide-iodine-malonic acid reaction (dimensionless) \\
I11-D1 & Naive critical slowing down (statistical mechanics) & I43-D2 & Driven pendulum with linear damping / Josephson junction (dimensionless) \\
I12-D1 & Photons in a laser (simple) & I44-D2 & Driven pendulum with quadratic damping (dimensionless) \\
I13-D1 & Overdamped bead on a rotating hoop & I45-D2 & Isothermal autocatalytic reaction model by Gray and Scott 1985 (dimensionless) \\
I14-D1 & Budworm outbreak model with predation & I46-D2 & Interacting bar magnets \\
I15-D1 & Budworm outbreak with predation (dimensionless) & I47-D2 & Binocular rivalry model (no oscillations) \\
I16-D1 & Landau equation (typical time scale tau = 1) & I48-D2 & Bacterial respiration model for nutrients and oxygen levels \\
I17-D1 & Logistic equation with harvesting/fishing & I49-D2 & Brusselator: hypothetical chemical oscillation model (dimensionless) \\
I18-D1 & Improved logistic equation with harvesting/fishing & I50-D2 & Chemical oscillator model by Schnackenberg 1979 (dimensionless) \\
I19-D1 & Improved logistic equation with harvesting/fishing (dimensionless) & I51-D2 & Oscillator death model by Ermentrout and Kopell 1990 \\
I20-D1 & Autocatalytic gene switching (dimensionless) & I52-D3 & Maxwell-Bloch equations (laser dynamics) \\
I21-D1 & Dimensionally reduced SIR infection model for dead people (dimensionless) & I53-D3 & Model for apoptosis (cell death) \\
I22-D1 & Hysteretic activation of a protein expression (positive feedback, basal promoter expression) & I54-D3 & Lorenz equations in well-behaved periodic regime \\
I23-D1 & Overdamped pendulum with constant driving torque/fireflies/Josephson junction (dimensionless) & I55-D3 & Lorenz equations in complex periodic regime \\
I24-D2 & Harmonic oscillator without damping & I56-D3 & Lorenz equations standard parameters (chaotic) \\
I25-D2 & Harmonic oscillator with damping & I57-D3 & Rössler attractor (stable fixed point) \\
I26-D2 & Lotka-Volterra competition model (Strogatz version with sheep and rabbits) & I58-D3 & Rössler attractor (periodic) \\
I27-D2 & Lotka-Volterra simple (as on Wikipedia) & I59-D3 & Rössler attractor (chaotic) \\
I28-D2 & Pendulum without friction & I60-D3 & Aizawa attractor (chaotic) \\
I29-D2 & Dipole fixed point & I61-D3 & Chen-Lee attractor; system for gyro motion with feedback control of rigid body (chaotic) \\
I30-D2 & RNA molecules catalyzing each others replication & I62-D4 & Binocular rivalry model with adaptation (oscillations) \\
I31-D2 & SIR infection model only for healthy and sick & I63-D4 & SEIR infection model (proportions) \\
I32-D2 & Damped double well oscillator &  &  \\
\bottomrule
    \end{tabular}
    \caption{Descriptions of the equations for the ODE datasets.}
    \label{tab:ode-dataset-description}
\end{table*}

\begin{table*}[ht]
\centering
\renewcommand{\arraystretch}{1.2}
\begin{tabular}{l|cccccccc}
\toprule
\textbf{Dataset} & SINDy & ESINDy & PySR & Operon & EQL & ODEFormer & End2end & uDSR \\
\midrule
I1-D1 & \checkmark & \checkmark & \checkmark & \checkmark  & - & \checkmark & $10^{2} (16)$ & \checkmark \\
I2-D1 & \checkmark & \checkmark & \checkmark & \underline{$10^{-9} (7)$}   & - & \checkmark & $10^{0} (27)$ & \underline{$10^{-9} (5)$} \\
I3-D1 & \checkmark & \underline{$10^{-5} (10)$} & $10^{0} (6)$ & $10^{0} (8)$   & - & \checkmark & $10^{2} (20)$ & $10^{2} (5)$ \\
I4-D1 & $10^{-2} (3)$ & $10^{-2} (3)$ & $10^{-2} (3)$ & $10^{-1} (7)$   & $10^{-2} (19)$ & $10^{-1} (9)$ & $10^{-1} (10)$ & $10^{-1} (5)$ \\
I5-D1 &  $10^{-1} (5)$ & $10^{-1} (5)$ & $10^{-1} (5)$ & $10^{-1} (7)$   & - & $10^{0} (5)$ & $10^{0} (27)$ & $10^{-1} (5)$ \\
I6-D1 & \checkmark & \checkmark & $10^{7} (6)$ & $10^{7} (8)$   & - & \checkmark & $10^{10} (65)$ & $10^{9} (6)$ \\
I7-D1 & $10^{-2} (3)$ & $10^{-2} (3)$ & $10^{-2} (3)$ & $10^{-4} (7)$   & $10^{1} (67)$ & $10^{-2} (3)$ & $10^{-2} (11)$ & $10^{-4} (5)$ \\
I8-D1 & $10^{-1} (5)$ & $10^{0} (1)$ & $10^{5} (6)$ & $10^{-1} (7)$   & - & $10^{0} (9)$ & $10^{0} (22)$ & $10^{-1} (5)$ \\
I9-D1 & \checkmark & \checkmark & \checkmark & \checkmark   & $10^{-2} (23)$ & \checkmark & $10^{3} (10)$ & \checkmark \\
I10-D1 & $10^{-2} (9)$ & $10^{-2} (9)$ & $10^{-2} (6)$ & $10^{0} (8)$   & $10^{1} (1)$ & $10^{-2} (9)$ & $10^{1} (16)$ & $10^{1} (7)$ \\
I11-D1 & $10^{0} (28)$ & \underline{$10^{-1} (5)$} & $10^{2} (10)$ & $10^{1} (10)$   & - & \checkmark & $10^{3} (30)$ & $10^{4} (5)$ \\
I12-D1 & $10^{2} (15)$ & \underline{$10^{3} (10)$} & $10^{7} (6)$ & $10^{6} (8)$   & - & $10^{7} (5)$ & $10^{10} (34)$ & $10^{8} (5)$ \\
I13-D1 & $10^{0} (12)$ & $10^{0} (12)$ & $10^{0} (7)$ & $10^{-1} (8)$   & $10^{1} (10)$ & $10^{1} (12)$ & $10^{3} (32)$ & $10^{3} (7)$ \\
I14-D1 & $10^{-4} (15)$ & $10^{1} (1)$ & $10^{-4} (7)$ & $10^{-1} (8)$   & - & $10^{-2} (9)$ & $10^{1} (36)$ & $10^{1} (7)$ \\
I15-D1 & $10^{0} (5)$ & $10^{0} (5)$ & - & $10^{-2} (8)$   & - & $10^{-1} (9)$ & $10^{1} (16)$ & $10^{0} (5)$ \\
I16-D1 & $10^{3} (10)$ & $10^{3} (10)$ & $10^{2} (8)$ & $10^{2} (13)$   & - & $10^{3} (18)$ & $10^{2} (25)$ & $10^{3} (7)$ \\
I17-D1 & $10^{0} (1)$ & $10^{0} (1)$ & \underline{$10^{-3} (7)$}& $10^{-3} (8)$   & - & \checkmark & $10^{0} (23)$ & $10^{0} (7)$ \\
I18-D1 & $10^{0} (5)$ & $10^{0} (5)$ & $10^{-4} (6)$ & $10^{-2} (8)$   & - & $10^{-1} (9)$ & $10^{0} (33)$ & $10^{0} (5)$ \\
I19-D1 & $10^{-1} (9)$ & $10^{-1} (9)$ & $10^{1} (5)$ & $10^{1} (8)$   & $10^{-2} (11)$ & $10^{1} (9)$ & $10^{4} (29)$ & $10^{3} (5)$ \\
I20-D1 & $10^{-4} (15)$ & $10^{0} (5)$ & $10^{-2} (6)$ & $10^{-1} (8)$   & $10^{1} (5)$ & $10^{-2} (19)$ & $10^{0} (11)$ & $10^{-1} (8)$ \\
I21-D1 & $10^{0} (1)$ & $10^{0} (1)$ & $10^{-2} (6)$ & $10^{-1} (8)$   & - & $10^{5} (23)$ & $10^{0} (30)$ & $10^{0} (5)$ \\
I22-D1 & $10^{-2} (18)$  & $10^{-1} (10)$  & $10^{1} (5)$ & $10^{1} (7)$   & $10^{0} (19)$ & $10^{1} (5)$ & $10^{1} (15)$ & $10^{1} (5)$ \\
I23-D1 & \checkmark & \checkmark & \checkmark & \checkmark   & $10^{0} (10)$ & $10^{3} (19)$ & $10^{5} (18)$ & $10^{-3} (6)$ \\
I24-D2 & \checkmark & \checkmark & \checkmark & \underline{$10^{-6} (14)$}   & $10^{-5} (18)$ & $10^{-2} (17)$ & $10^{0} (72)$ & \checkmark \\
I25-D2 & \checkmark & \checkmark & \checkmark & \underline{$10^{-5} (18)$}   & $10^{2} (112)$ & $10^{-3} (11)$ & $10^{0} (58)$ & $10^{-4} (14)$ \\
I26-D2 & $10^{-1} (48)$ & $10^{4} (36)$ & $10^{0} (12)$ & $10^{3} (19)$   & $10^{1} (90)$ & $10^{4} (35)$ & $10^{4} (114)$ & $10^{4} (18)$ \\
I27-D2 & $10^{3} (19)$ & $10^{0} (10)$ & \checkmark & $10^{-3} (27)$   & - & $10^{-1} (19)$ & $10^{0} (67)$ & $10^{-1} (13)$ \\
I28-D2 & \checkmark & \checkmark & \checkmark & \underline{$10^{-6} (15)$}   & $10^{-2} (16)$ & $10^{-1} (25)$ & $10^{0} (76)$ & $10^{-2} (11)$ \\
I29-D2 & $10^{-2} (33)$ & $10^{-1} (28)$ & \checkmark & $10^{2} (23)$   & $10^{0} (102)$ & $10^{-1} (21)$ & $10^{2} (63)$ & $10^{1} (12)$ \\
I30-D2 & $10^{-1} (6)$ & $10^{-1} (6)$ & $10^{-1} (6)$ & $10^{-1} (14)$   & $10^{-1} (2)$ & $10^{-1} (11)$ & $10^{0} (53)$ & $10^{-1} (9)$ \\
I31-D2 & $10^{-2} (21)$ & $10^{-3} (21)$ & \checkmark & $10^{-3} (20)$   & - & \underline{$10^{-8} (19)$} & $10^{1} (57)$ & $10^{-1} (19)$ \\
I32-D2 & $10^{-1} (11)$ & $10^{-1} (11)$ & \underline{ $10^{-2} (8)$} & $10^{-2} (19)$   & $10^{-1} (34)$ & $10^{0} (23)$ & $10^{0} (59)$ & $10^{-1} (13)$ \\
I33-D2 & $10^{0} (10)$ & $10^{0} (10)$ & $10^{0} (14)$ & $10^{0} (20)$   & $10^{0} (16)$ & $10^{0} (9)$ & $10^{1} (106)$ & $10^{0} (17)$ \\
I34-D2 & $10^{-2} (6)$ & $10^{-2} (6)$ & $10^{-2} (7)$ & $10^{-4} (20)$   & $10^{-2} (77)$ & $10^{-2} (37)$ & $10^{0} (92)$ & $10^{-2} (8)$ \\
I35-D2 & $10^{-1} (17)$ & $10^{-1} (17)$ & $10^{-2} (16)$ & $10^{-2} (31)$   & $10^{-2} (98)$ & $10^{-1} (21)$ & - & $10^{-1} (12)$ \\

\bottomrule
\end{tabular}
\caption{Performance of MD methods on ODE datasets, measured by the NMSE of the time derivatives. Reported NMSEs are rounded to the nearest power of 10. Equation complexity is given in parenthesis. A check mark (\checkmark) indicates successful identification of the full equation, with all coefficients being in $\pm5\%$ of ground truth. Underlined entries denote partially correct equations, where only one term is missing or extra, and the remaining terms have coefficients that are within $\pm5\%$ of ground truth. Empty entries (-) show method failure or timeout.}
\label{tab:ode-performance-1}
\end{table*}

\begin{table*}[ht]
\centering
\renewcommand{\arraystretch}{1.2}
\begin{tabular}{l|cccccccc}
\toprule
\textbf{Dataset} & SINDy & ESINDy & PySR & Operon & EQL & ODEFormer & End2end & uDSR \\
\midrule

I36-D2 & $10^{-5} (11)$ & $10^{-2} (10)$ & $10^{-2} (6)$ & $10^{-1} (18)$ & $10^{-1} (16)$ & $10^{-2} (11)$ & $10^{2} (56)$ & $10^{-3} (13)$ \\
I37-D2 & $10^{-2} (15)$ & $10^{-2} (25)$ & \checkmark & $10^{-2} (23)$  & $10^{-1} (42)$ & $10^{-1} (33)$ & $10^{0} (83)$ & $10^{-1} (14)$ \\
I38-D2 & \underline{$10^{-3} (20)$} & \underline{$10^{-3} (20)$} & $10^{-2} (15)$ & $10^{-3} (25)$  & $10^{0} (152)$ & $10^{0} (32)$ & $10^{0} (83)$ & $10^{0} (15)$ \\
I39-D2 & $10^{-2} (12)$ & $10^{-2} (12)$ & $10^{-3} (10)$ & $10^{-2} (18)$  & $10^{-1} (38)$ & $10^{-1} (13)$ & $10^{1} (42)$ & $10^{-2} (13)$ \\
I40-D2 & $10^{-1} (6)$ & $10^{0} (15)$ & $10^{-2} (10)$ & $10^{-4} (26)$  & $10^{-1} (38)$ & $10^{0} (26)$ & $10^{0} (54)$ & $10^{0} (15)$ \\
I41-D2 & $10^{-1} (13)$ & $10^{-4} (56)$ & $10^{-1} (17)$ & $10^{-1} (23)$  & $10^{0} (122)$ & $10^{2} (32)$ & $10^{0} (134)$ & $10^{1} (12)$ \\
I42-D2 & $10^{-1} (13)$ & $10^{-1} (13)$ & $10^{-2} (23)$ & $10^{-2} (38)$  & $10^{-1} (64)$ & $10^{-1} (21)$ & $10^{0} (94)$ & $10^{-1} (14)$ \\
I43-D2 & \checkmark & \checkmark & \checkmark & $10^{-3} (19)$  & - & $10^{-1} (18)$ & $10^{0} (65)$ & $10^{-3} (17)$ \\
I44-D2 & $10^{-5} (18)$ & $10^{-3} (12)$ & $10^{-4} (12)$ & $10^{-1} (19)$  & - & $10^{-1} (22)$ & $10^{0} (65)$ & $10^{-5} (17)$ \\
I45-D2 & $10^{-1} (16)$ & $10^{0} (16)$ & \underline{$10^{-2} (14)$} & $10^{-1} (20)$  & $10^{-1} (42)$ & $10^{-2} (23)$ & $10^{0} (95)$ & $10^{-1} (15)$ \\
I46-D2 & $10^{-3} (14)$ & $10^{-4} (19)$ & $10^{-1} (9)$ & $10^{0} (18)$  & $10^{2} (2)$ & $10^{-2} (17)$ & $10^{2} (62)$ & $10^{-2} (17)$ \\
I47-D2 & $10^{-2} (16)$ & $10^{-2} (16)$ & $10^{0} (10)$ & $10^{-1} (24)$  & $10^{1} (28)$ & $10^{1} (27)$ & $10^{1} (73)$ & $10^{0} (10)$ \\
I48-D2 & $10^{0} (16)$ & $10^{-1} (16)$ & $10^{0} (27)$ & $10^{-1} (34)$  & - & $10^{1} (27)$ & $10^{0} (78)$ & $10^{-1} (16)$ \\
I49-D2 & $10^{0} (16)$ & $10^{2} (76)$ & \checkmark & \underline{$10^{-3} (36)$}  & $10^{1} (52)$ & $10^{1} (25)$ & $10^{1} (63)$ & $10^{0} (15)$ \\
I50-D2 & $10^{3} (68)$ & $10^{2} (36)$ & $10^{-2} (21)$ & \checkmark  & $10^{2} (152)$ & $10^{5} (33)$ & $10^{3} (56)$ & $10^{4} (19)$ \\
I51-D2 & $10^{0} (16)$ & $10^{-1} (16)$ & $10^{-1} (18)$ & $10^{0} (26)$  & - & $10^{-1} (24)$ & $10^{-1} (109)$ & $10^{-1} (17)$ \\
I52-D3 & $10^{-1} (29)$ & $10^{-1} (29)$ & $10^{-1} (20)$ & $10^{-1} (28)$  & $10^{0} (181)$ & $10^{2} (25)$ & $10^{0} (138)$ & $10^{-1} (28)$ \\
I53-D3 & $10^{-1} (57)$ & - & $10^{3} (23)$ & $10^{3} (31)$  & $10^{2} (45)$ & $10^{3} (33)$ & $10^{5} (105)$ & $10^{3} (31)$ \\
I54-D3 & $10^{-1} (23)$ & $10^{-1} (23)$ & \checkmark & $10^{-2} (27)$  & $10^{-2} (213)$ & $10^{0} (35)$ & $10^{0} (120)$ & $10^{-1} (37)$ \\
I55-D3 & $10^{0} (26)$ & $10^{0} (26)$ & $10^{-1} (52)$& $10^{-1} (75)$  & - & $10^{-1} (34)$ & $10^{0} (168)$ & $10^{0} (43)$ \\
I56-D3 & $10^{-1} (52)$ & $10^{-1} (47)$ & $10^{-2} (22)$ & $10^{-1} (34)$  & - & $10^{0} (46)$ & $10^{0} (120)$ & $10^{-1} (36)$ \\
I57-D3 & $10^{-2} (19)$ & $10^{-1} (19)$ & $10^{-3} (17)$ & $10^{-1} (26)$  & $10^{-2} (93)$ & $10^{0} (21)$ & $10^{1} (115)$ & $10^{-1} (32)$ \\
I58-D3 & $10^{-1} (24)$ & - & $10^{-2} (21)$ & $10^{-2} (28)$  & - & $10^{0} (29)$ & $10^{0} (134)$ & $10^{-1} (29)$ \\
I59-D3 & $10^{-1} (13)$ & $10^{-1} (17)$ & $10^{-1} (23)$ & $10^{-2} (39)$  & - & $10^{0} (29)$ & $10^{0} (169)$ & $10^{0} (33)$ \\
I60-D3 & $10^{-1} (18)$  & - & $10^{-3} (34)$ & $10^{-1} (42)$   & $10^{-1} (39)$ & $10^{-1} (25)$ & $10^{0} (90)$ & $10^{-1} (30)$ \\
I61-D3 & $10^{-1} (27)$ & $10^{-1} (23)$ & $10^{-2} (35)$ & $10^{-2} (56)$  & - & $10^{-1} (33)$ & $10^{0} (132)$ & $10^{-1} (25)$ \\
I62-D4 & $10^{1} (27)$ & $10^{1} (27)$ & $10^{2} (14)$ & $10^{3} (30)$  & $10^{1} (216)$ & $10^{2} (37)$ & $10^{4} (127)$ & $10^{2} (24)$ \\
I63-D4 & $10^{-3} (33)$ & $10^{-3} (33)$ & $10^{-5} (20)$ & $10^{-2} (35)$  & $10^{0} (4)$ & $10^{1} (41)$ & $10^{1} (138)$ & $10^{-1} (28)$ \\

\bottomrule
\end{tabular}
\caption{Performance of MD methods on ODE datasets, measured by the NMSE of the time derivatives. Reported NMSEs are rounded to the nearest power of 10. Equation complexity is given in parenthesis. A check mark (\checkmark) indicates successful identification of the full equation, with all coefficients being in $\pm5\%$ of ground truth. Underlined entries denote partially correct equations, where only one term is missing or extra, and the remaining terms have coefficients that are within $\pm5\%$ of ground truth. Empty entries (-) show method failure or timeout.}
\label{tab:ode-performance-2}
\end{table*}

\begin{table*}
    \centering
\begin{tabular}{lllllllll}
\toprule
Dataset & PDEFind & Bayesian & WSINDy & EWSINDy & DeepMoD & PySR & Operon & uDSR \\
\midrule
Advection & 156 & 4773 & 43 & 6600 & 21194 & 60 & 345 & 302 \\
Burgers & 25 & 40 & 25 & 987 & 1528 & 30 & 24 & 152 \\
KdV & 58 & 332 & 31 & 3100 & 2786 & 44 & 265 & 220 \\
KS & 82 & 968 & 49 & 6099 & 10073 & 93 & 799 & 200 \\
AD & 664 & 33803 & 196 & 3846 & 2294 & 76 & 687 & 183 \\
Heat (Soil) 1D & 21 & 56 & 24 & 897 & 1172 & 31 & 17 & 110 \\
Heat (Soil) 2D & 835 & - & 147 & 7125 & - & 1042 & 11841 & 98 \\
Heat (Soil) 3D & 9760 & - & - & - & - & 385 & 2704 & 224 \\
Heat (Laser) & 195 & - & - & - & - & 2664 & 22282 & 462 \\
RD & 7751 & - & 1204 & - & - & 132 & 604 & 2138 \\
NLS & 466 & - & 326 & - & - & 59 & 204 & 288 \\
NS (Channel) & 9 & - & - & - & - & 52 & 5 & 2040 \\
NS (Cylinder) & 278 & - & 119 & 3328 & - & 636 & 1999 & 640 \\
RD (Cylinder) & - & - & 975 & 17218 & - & 2026 & 3772 & 742 \\
\bottomrule
\end{tabular}
    \caption{The runtime of each algorithm on clean PDE datasets in seconds. }
    \label{tab:pde-times}
\end{table*}

\clearpage

\bibliography{aaai2026-refs}

\clearpage

\makeatletter
\@ifundefined{isChecklistMainFile}{
  \newif\ifreproStandalone
  \reproStandalonetrue
}{
  \newif\ifreproStandalone
  \reproStandalonefalse
}
\makeatother

\ifreproStandalone

\fi
\setlength{\leftmargini}{20pt}
\makeatletter\def\@listi{\leftmargin\leftmargini \topsep .5em \parsep .5em \itemsep .5em}
\def\@listii{\leftmargin\leftmarginii \labelwidth\leftmarginii \advance\labelwidth-\labelsep \topsep .4em \parsep .4em \itemsep .4em}
\def\@listiii{\leftmargin\leftmarginiii \labelwidth\leftmarginiii \advance\labelwidth-\labelsep \topsep .4em \parsep .4em \itemsep .4em}\makeatother

\setcounter{secnumdepth}{0}
\renewcommand\thesubsection{\arabic{subsection}}
\renewcommand\labelenumi{\thesubsection.\arabic{enumi}}

\newcounter{checksubsection}
\newcounter{checkitem}[checksubsection]

\newcommand{\checksubsection}[1]{%
  \refstepcounter{checksubsection}%
  \paragraph{\arabic{checksubsection}. #1}%
  \setcounter{checkitem}{0}%
}

\newcommand{\checkitem}{%
  \refstepcounter{checkitem}%
  \item[\arabic{checksubsection}.\arabic{checkitem}.]%
}
\newcommand{\question}[2]{\normalcolor\checkitem #1 #2 \color{blue}}
\newcommand{\ifyespoints}[1]{\makebox[0pt][l]{\hspace{-15pt}\normalcolor #1}}

\end{document}